\documentclass[10pt,twocolumn,letterpaper]{article}

\usepackage[pagenumbers]{iccv} 

\usepackage[accsupp]{axessibility}

\definecolor{iccvblue}{rgb}{0.21,0.49,0.74}
\usepackage[pagebackref,breaklinks,colorlinks,allcolors=iccvblue]{hyperref}

\usepackage{adjustbox}
\usepackage{multirow}
\usepackage{bm}

\title{ZIM: Zero-Shot Image Matting for Anything}

\author{
Beomyoung Kim
~~~~~~~~Chanyong Shin
~~~~~~~~Joonhyun Jeong
~~~~~~~~Hyungsik Jung\\
~~~~~~~~Se-Yun Lee
~~~~~~~~Sewhan Chun
~~~~~~~~Dong-Hyun Hwang
~~~~~~~~Joonsang Yu \\[1.2mm]
{NAVER Cloud, ImageVision}\\
}

\begin{document}

\twocolumn[{%
\renewcommand\twocolumn[1][]{#1}%
\maketitle
\begin{center}
    \vspace{-5mm}
    \includegraphics[width=0.98\linewidth]{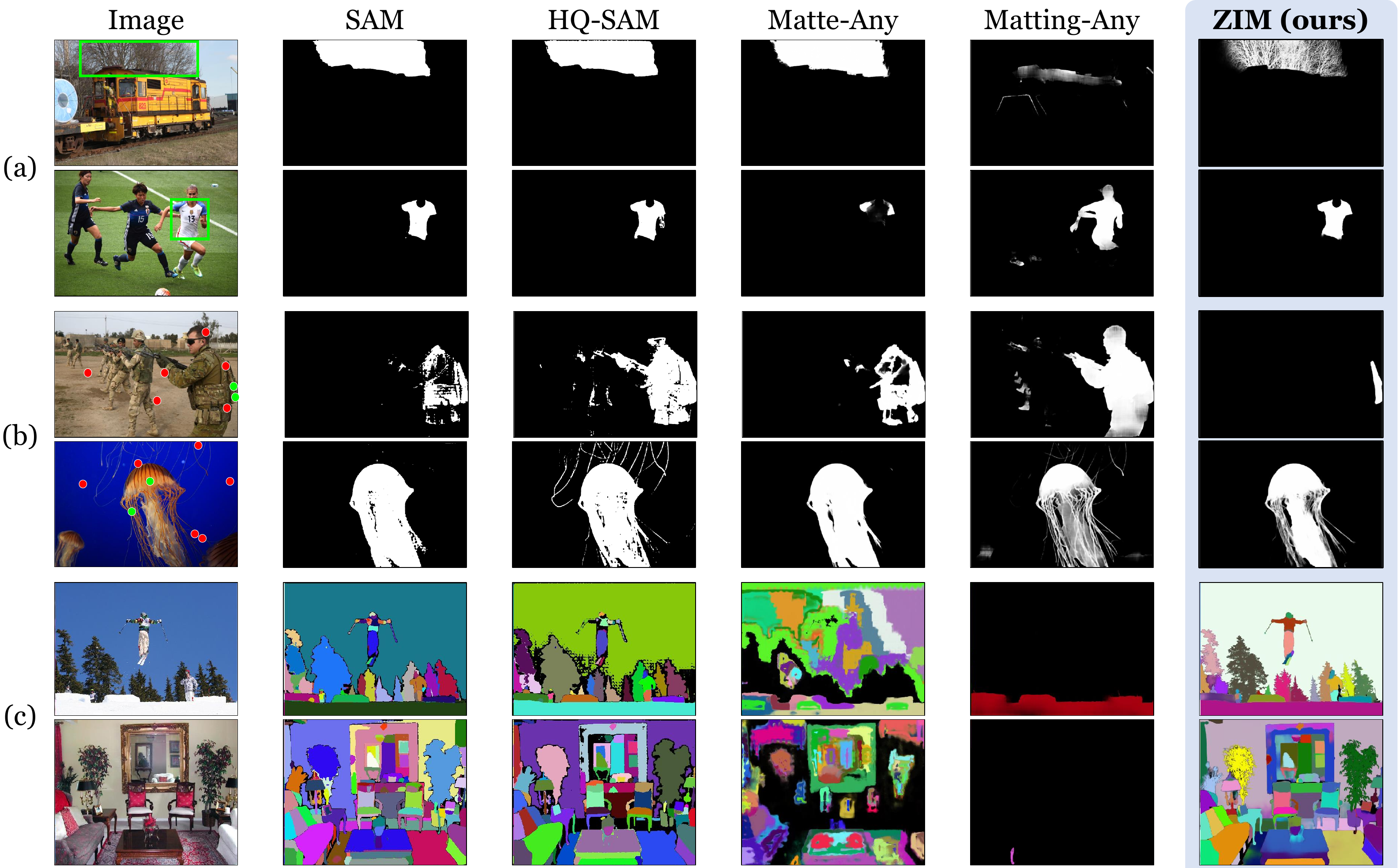}
    \vspace{-1mm}
    \captionof{figure}{\textbf{Qualitative comparison} of ours with five existing zero-shot models (SAM~\cite{(SAM)kirillov2023segment}, HQ-SAM~\cite{(HQ-SAM)ke2024segment}, Matte-Any~\cite{(matte-any)yao2024matte}, and Matting-Any~\cite{(matting-any)li2024matting}). 
    It showcases (a) box prompting results, (b) point prompting results, and (c) automatic mask generation results.
    }
    \label{fig:cover}
\end{center}
}]

\begin{abstract}

The recent segmentation foundation model, Segment Anything Model (SAM), exhibits strong zero-shot segmentation capabilities, but it falls short in generating fine-grained precise masks.
To address this limitation, we propose a novel zero-shot image matting model, called ZIM, with two key contributions: First, we develop a label converter that transforms segmentation labels into detailed matte labels, constructing the new SA1B-Matte dataset without costly manual annotations. Training SAM with this dataset enables it to generate precise matte masks while maintaining its zero-shot capability. Second, we design the zero-shot matting model equipped with a hierarchical pixel decoder to enhance mask representation, along with a prompt-aware masked attention mechanism to improve performance by enabling the model to focus on regions specified by visual prompts.
We evaluate ZIM using the newly introduced MicroMat-3K test set, which contains high-quality micro-level matte labels. 
Experimental results show that ZIM outperforms existing methods in fine-grained mask generation and zero-shot generalization. 
Furthermore, we demonstrate the versatility of ZIM in various downstream tasks requiring precise masks, such as image inpainting and 3D segmentation.
Our contributions provide a robust foundation for advancing zero-shot matting and its downstream applications across a wide range of computer vision tasks. 
The code is available at \url{https://naver-ai.github.io/ZIM}.
\enlargethispage{\baselineskip}

\end{abstract}

\section{Introduction}

Image segmentation, which divides an image into distinct regions to facilitate subsequent analysis, is a fundamental task in computer vision.
Recent breakthroughs in segmentation models have made significant strides in this area, particularly with the emergence of the segmentation foundation model, Segment Anything Model (SAM)~\cite{(SAM)kirillov2023segment}.
SAM is trained on the SA1B dataset~\cite{(SAM)kirillov2023segment} containing 1 billion micro-level segmentation labels, where its extensiveness enables SAM to generalize effectively across a broad range of tasks.
Its strong zero-shot capabilities, powered by visual prompts, have redefined the state of the art in zero-shot interactive segmentation and opened new avenues for tackling more complex tasks within the zero-shot paradigm.

\begin{figure*}[t]
    \centering
    \includegraphics[width=0.98\linewidth]{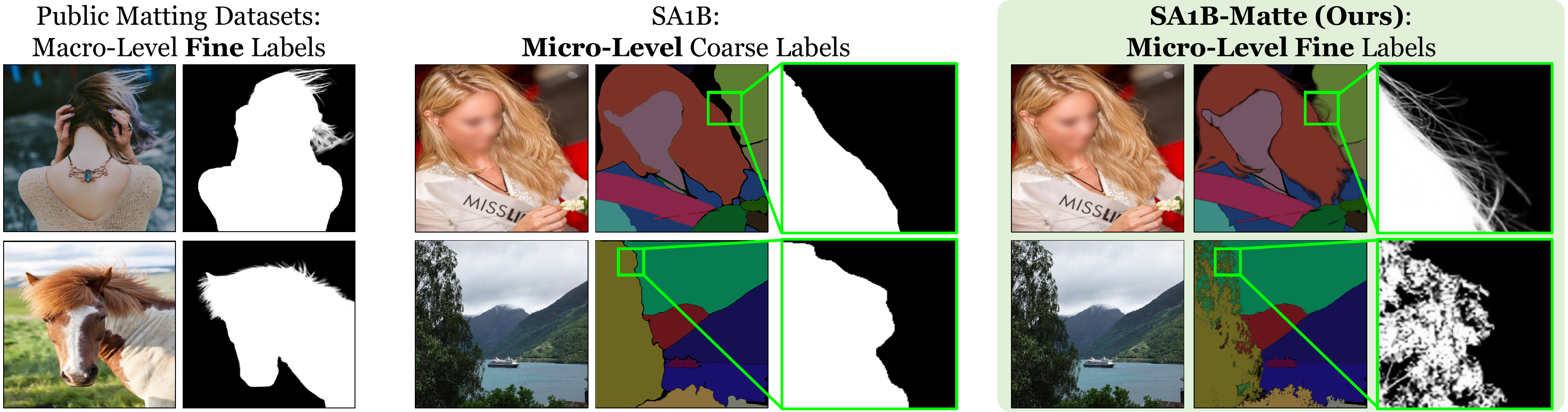}
    \vspace{-1mm}
    \caption{
        \textbf{Qualitative samples from each dataset:} Public matting datasets~\cite{(AIM-500)li2021deep,(AM-2K)li2022bridging,(D646)qiao2020attention,(composition-1k)xu2017deep} with macro-level fine labels, the SA1B datset~\cite{(SAM)kirillov2023segment} with micro-level coarse labels, and our proposed SA1B-Matte dataset incorporating the micro-level labels with fine details.
    }
    \vspace{-2mm}
    \label{fig:dataset}
\end{figure*}

Despite these achievements, SAM often struggles to generate masks with fine-grained precision (see Figure \ref{fig:cover}). To address this limitation, recent studies~\cite{(matte-any)yao2024matte, (matting-any)li2024matting, (smat)ye2024unifying} have extended SAM to the image matting task, which focuses on capturing highly detailed boundaries and intricate details such as individual hair strands. These approaches achieve enhanced mask precision by fine-tuning SAM on publicly available matting datasets~\cite{(AIM-500)li2021deep,(D646)qiao2020attention,(composition-1k)xu2017deep}. However, this fine-tuning process can undermine the zero-shot potential of SAM, since most public matting datasets contain only macro-level labels ($e.g.,$ entire human portrait) rather than the more detailed micro-level labels ($e.g.,$ individual body parts), as illustrated in Figure \ref{fig:dataset}. Fine-tuning with macro-level labels can cause SAM to overfit to this macro-level granularity, resulting in catastrophic forgetting of its ability to generalize at the micro-level granularity, as shown in Figure \ref{fig:cover}. Moreover, the scarcity of large-scale matting datasets with micro-level matte labels poses a significant obstacle in developing effective zero-shot matting solutions.

In this paper, we introduce a pioneering \textbf{Z}ero-shot \textbf{I}mage \textbf{M}atting model, dubbed \textbf{ZIM}, that retains strong zero-shot capabilities while generating high-quality micro-level matting masks. 
A key challenge in this domain is the need for a matting dataset with extensive micro-level matte labels, which are costly and labor-intensive to annotate. 
To address this challenge, we propose a novel label conversion method that transforms any segmentation label into a detailed matte label. 
For more reliable label transformation, we design two effective strategies to reduce noise and yield high-fidelity matte labels (Section \ref{section3.1}).
Subsequently, we construct a new dataset, called \textbf{SA1B-Matte}, which contains an extensive set of micro-level matte labels generated by transforming segmentation labels from the SA1B dataset via the proposed converter (see Figure \ref{fig:dataset}).
By training SAM on the SA1B-Matte dataset, we introduce an effective foundational matting model with micro-level granularity while preserving the zero-shot ability of SAM (see Figure \ref{fig:cover}).

To further ensure effective image matting, we enhance the major bottleneck in the network architecture of SAM that impedes capturing robust and detailed feature maps. Specifically, SAM employs a simple pixel decoder to generate mask feature maps with a stride of 4, which is susceptible to checkerboard artifacts and often falls short in capturing fine details.
To mitigate this, we design a more elaborated pixel decoder, enabling more robust and richer mask representations (Section \ref{section3.2}).
Furthermore, we introduce a prompt-aware masked attention mechanism that leads to the improvement of interactive matting performance. 

To validate our zero-shot matting model, we present a new test set, called \textbf{MicroMat-3K}, consisting of 3,000 high-quality micro-level matte labels. Our experiments on this dataset demonstrate that while SAM exhibits strong zero-shot capabilities, it struggles to deliver precise mask outputs. In contrast, existing matting models show limited zero-shot performance. ZIM, however, not only maintains robust zero-shot functionality but also provides superior precision in mask generation. Additionally, we highlight the foundational applicability of ZIM in several downstream tasks requiring precise masks, such as image inpainting~\cite{(inpaint_anything)yu2023inpaint} and 3D segmentation~\cite{(SA3D)cen2023segment}. We hope this work provides valuable insights to the research community, encouraging further development of zero-shot matting models.

\section{Related Work}

\vspace{0.75mm} \noindent \textbf{Image Segmentation.}
Image segmentation is a fundamental task in computer vision, enabling the division of an image into distinct regions.
Recent advancements in segmentation models~\cite{(mask2former)cheng2022masked,(OneFormer)jain2023oneformer,(MaskDino)li2023mask} have significantly improved the accuracy of segmentation tasks, including semantic and instance segmentation.
The emergence of Segment Anything Model (SAM)~\cite{(SAM)kirillov2023segment} introduced a new paradigm in segmentation by leveraging visual prompts ($e.g.,$ points or boxes).
SAM is designed as a foundational segmentation model capable of handling diverse tasks due to its robust zero-shot capabilities, showing remarkable versatility across a wide range of tasks and domains. 
However, despite its strengths, SAM struggles to produce high-precision masks.
In this paper, we address this limitation by developing a novel zero-shot model that enhances mask precision while maintaining SAM’s generalization capabilities.

\vspace{0.75mm} \noindent \textbf{Image Matting.}
Image matting is a more complex task than image segmentation, as it focuses on estimating the soft transparency of object boundaries to capture fine details, which is critical in tasks like image compositing and background removal.
Unlike segmentation, which assigns hard labels to each pixel, matting requires precise edge detection and soft labeling for smooth blending between objects and their background.
Recent developments in zero-shot matting have aimed to build upon the foundational segmentation capabilities of SAM. 
Most approaches~\cite{(matting-any)li2024matting,(matte-any)yao2024matte,(smat)ye2024unifying} fine-tune SAM on public matting datasets~\cite{(AIM-500)li2021deep,(AM-2K)li2022bridging,(D646)qiao2020attention,(composition-1k)xu2017deep}. However, these datasets predominantly contain macro-level labels, degrading SAM's ability to generalize on micro-level structures, such as individual body parts of a human. The reliance on these datasets can deteriorate the zero-shot generalization of the model. Furthermore, the lack of large-scale matting datasets with micro-level labels restricts progress in developing matting models with truly effective zero-shot ability.
In this paper, we correspondingly construct a large-scale micro-level labeled matting dataset via our proposed label converter without laborious annotation procedures, enabling effective zero-shot matting modeling.

\begin{figure*}[t]
    \centering
        \begin{subfigure}[b]{0.40\linewidth}
        \centering
        \includegraphics[width=\linewidth]{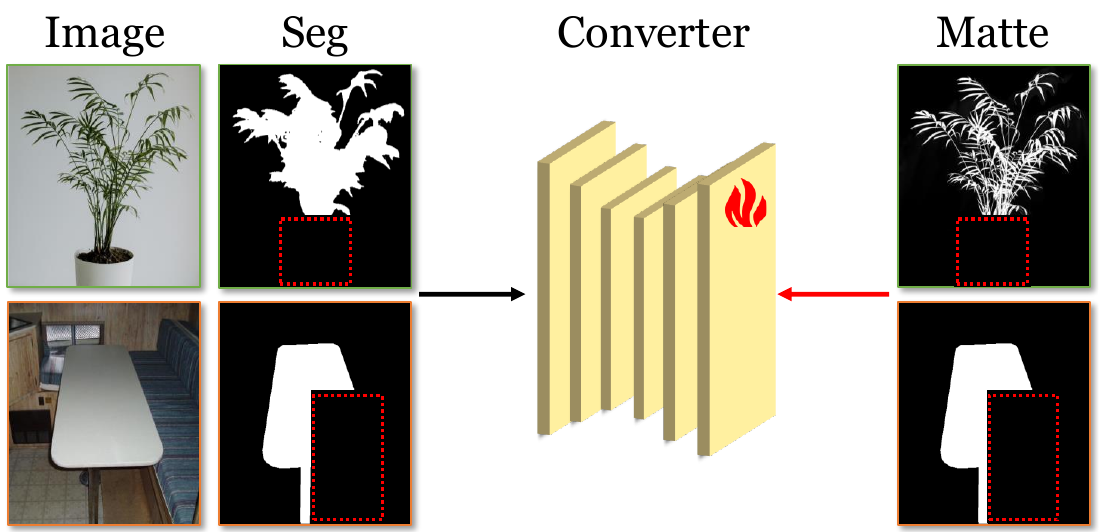} 
        \caption{Spatial Generalization Augmentation}
        \label{fig:converter_training} 
    \end{subfigure}
    \hspace{1mm}
    \begin{subfigure}[b]{0.57\linewidth}
        \centering
        \includegraphics[width=\linewidth]{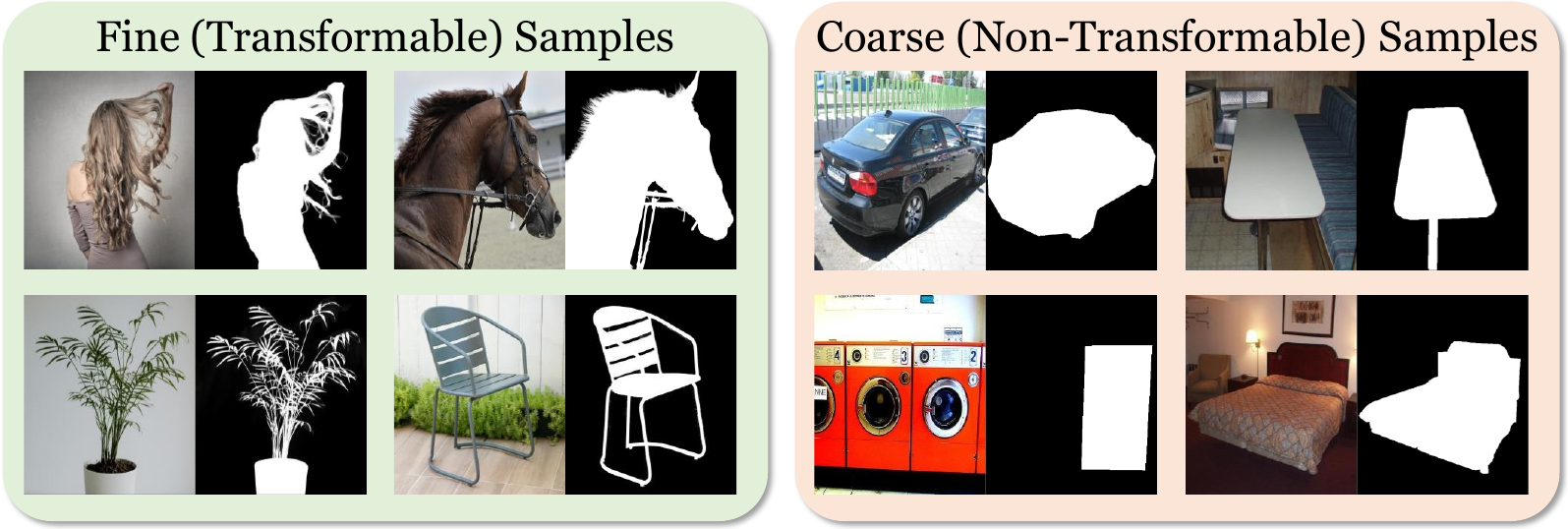} 
        \caption{Selective Transformation Learning}
        \label{fig:converter_label}  
    \end{subfigure}
    \vspace{-1mm}
    \caption{
    \textbf{Illustration of the key components of the Label Converter}.
        (a) Overview of the training procedure of the converter using Spatial Generalization Augmentation (indicated by red dotted boxes).
        (b) Examples of transformable (fine) and non-transformable (coarse) samples used in Selective Transformation Learning for the converter.
    }
    \label{fig:converter}
    \vspace{-1mm}
\end{figure*}

\section{Methodology}

Our contributions can be divided into two components: matting dataset construction (Section \ref{section3.1}) and network architecture enhancements (Section \ref{section3.2}).

\subsection{Constructing the Zero-Shot Matting Dataset}
\label{section3.1}

\noindent \textbf{Motivation.}
For effective zero-shot matting, a dataset with micro-level matte labels is essential.
However, manually annotating matte labels at the micro-level requires extensive human labor and cost.
To this end, we present an innovative \textbf{Label Converter} that transforms any segment label into a matte label, motivated by mask-guided matting works~\cite{(MGM_wild)park2023mask,(MGM)(RWP-636)yu2021mask}.
We first collect public image matting datasets~\cite{(AIM-500)li2021deep,(AM-2K)li2022bridging,(P3M)li2021privacy,(MGM)(RWP-636)yu2021mask,(HIM-2K)sun2022human,(RefMatte)li2023referring} to train the converter.
We derive coarse segmentation labels from matte labels by applying image processing techniques such as thresholding, resolution down-scaling, Gaussian blurring, dilation, erosion, and convex hull transformations.
The converter takes an image and segmentation label as input source and is trained to produce a corresponding matte label, as illustrated in Figure \ref{fig:converter_training}.

\vspace{0.75mm} \noindent \textbf{Challenges.}
\textbf{(1)} Generalization to unseen patterns: Public matting datasets predominantly contain macro-level labels ($e.g.,$ entire portraits), as shown in Figure \ref{fig:dataset}. Consequently, the converter trained on these datasets often struggles to generalize to unseen micro-level objects ($e.g.,$ individual body parts). This limitation leads to the generation of noisy matte labels when applied to micro-level segmentation (see the 4th column in Figure \ref{fig:analysis_SGA}).
\textbf{(2)} Unnecessary fine-grained representation: Some objects, such as cars or boxes, commonly do not require fine-grained representation. However, since the converter is trained to always transform segmentation labels into fine-grained matte labels, it often generates unnecessary noise into the output matte, particularly for objects that do not benefit from fine-grained representation (see the 4th column in Figure \ref{fig:analysis_STL}). 

\vspace{0.75mm} \noindent \textbf{Spatial Generalization Augmentation.}
To improve the converter’s ability to generalize to diverse segmentation labels, we design Spatial Generalization Augmentation. This approach introduces variability into the training data by applying a random cut-out technique, as shown in Figure \ref{fig:converter_training}. During training, both the segmentation label and the corresponding matte label are randomly cropped in the same regions. By exposing the converter to irregular and incomplete input patterns, this augmentation forces the converter to adapt to diverse spatial structures and unseen patterns, thus enhancing its generalization capability. This method ensures that the converter can better handle a variety of input segmentation labels, even those that deviate from training patterns (see the 3rd column in Figure \ref{fig:analysis_SGA}).

\vspace{0.75mm} \noindent \textbf{Selective Transformation Learning.}
To prevent the unnecessary transformation of objects that do not require fine-grained details ($e.g.,$ cars or desks), we introduce Selective Transformation Learning. 
This technique enables the converter to selectively focus on objects requiring detailed matte conversion ($e.g.,$ hair, trees) while skipping finer transformations for coarse-grained objects. 
We incorporate these non-transformable samples into the training process by collecting coarse-grained object masks from public segmentation datasets~\cite{(ade20k)zhou2017scene} (see Figure \ref{fig:converter_label}).
During training, the ground-truth matte label for the non-transformable samples is set to identical to the original segmentation label, allowing the converter to learn that no transformation is required. 
This selective approach reduces noise in the output and ensures that fine-grained transformations are applied only when needed (see the 3rd column in Figure \ref{fig:analysis_STL}).

\vspace{0.75mm} \noindent \textbf{Training.}
We employ standard loss functions commonly used in matting tasks, namely using a linear combination of L1 and Gradient losses~\cite{(PPM)ke2022modnet,(AIM-500)li2021deep,(AM-2K)li2022bridging} to minimize pixel-wise differences between the ground-truth and predicted matte:
\begin{gather}
    L = L_{l1} + \lambda L_{grad} \label{eq:loss_function}\\
    L_{l1} = |M - M'| \\
    L_{grad} = |\nabla_{x}(M) {-} \nabla_{x}(M')| + |\nabla_{y}(M) {-} \nabla_{y}(M')|
\end{gather}
where $M$ and $M'$ represent the ground-truth and predicted matte label, respectively, and $\lambda$ is a loss weighting factor.
In addition, $\nabla_{x}$ and $\nabla_{y}$ represent the gradients along the horizontal and vertical axes, respectively.
Moreover, we set a probability parameter $p$ to control the random application of Spatial Generalization Augmentation during training.

\begin{figure*}[t]
    \centering
    \includegraphics[width=\linewidth]{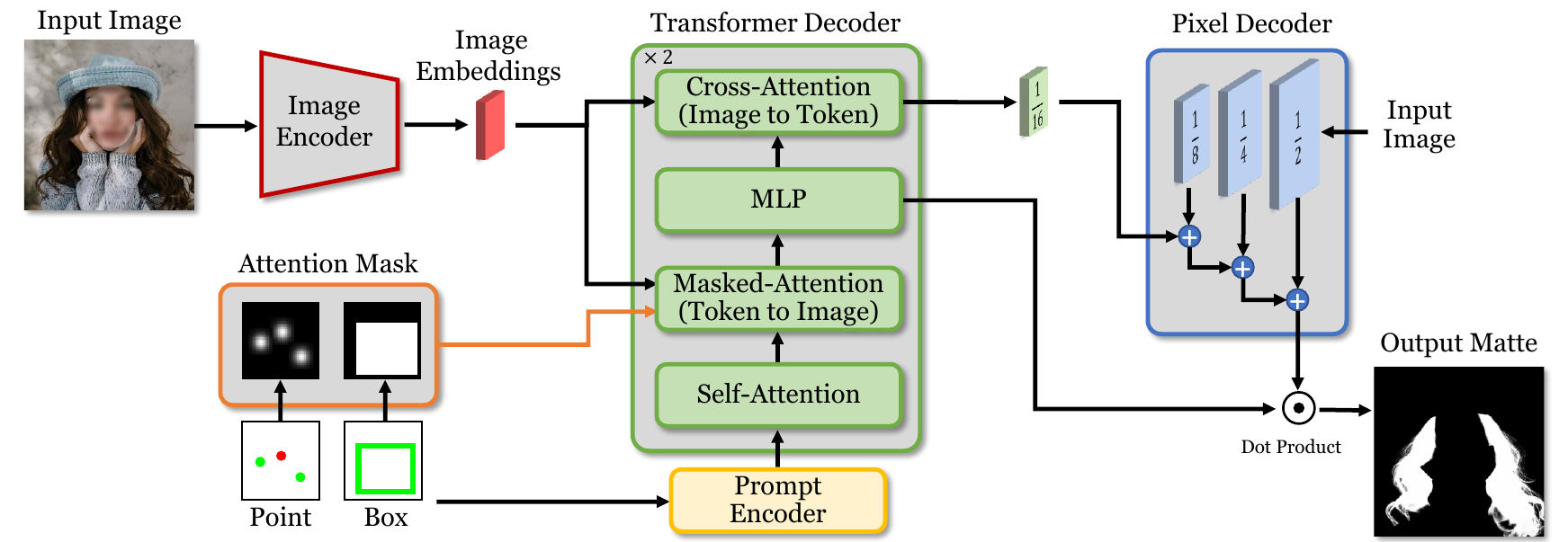}
    \caption{\textbf{Overview of the ZIM architecture.} 
    Based on the SAM network architecture~\cite{(SAM)kirillov2023segment}, we introduce two key improvements: (1) Hierarchical Pixel Decoder for more robust and higher-resolution mask feature map generation, and (2) Prompt-Aware Masked Attention mechanism to enhance interactive matting performance.} 
    \label{fig:overview}
    \vspace{-2mm}
\end{figure*}

\vspace{0.75mm} \noindent \textbf{SA1B-Matte Dataset.}
After training the label converter, we transform segmentation labels in the SA1B dataset~\cite{(SAM)kirillov2023segment} to matte labels using the converter, constructing a new SA1B-Matte dataset.
As shown in Figure \ref{fig:dataset}, the coarse labels in the SA1B dataset are successfully transformed into high-quality precise matte labels.
Compared to existing public matting datasets consisting of macro-level fine labels, the SA1B-Matte dataset is a large-scale image matting dataset with micro-level fine labels, providing an ideal foundation for developing zero-shot matting models.

\subsection{ZIM: Zero-Shot Image Matting Model}
\label{section3.2}

\vspace{0.75mm} \noindent \textbf{Overview of ZIM.}
Our proposed model, ZIM, builds upon SAM~\cite{(SAM)kirillov2023segment} and consists of four components, as illustrated in Figure \ref{fig:overview}:
(1) Image Encoder: extracts image features from the input image, producing an image embedding with a stride of 16.
(2) Prompt Encoder: encodes point or box inputs into prompt embeddings concatenated with learnable token embeddings, serving a role similar to the \texttt{[cls]} token in ViT~\cite{(vit)dosovitskiy2020image}.
(3) Transformer Decoder: takes the image and token embeddings to generate output token embeddings. It performs four operations: self-attention on the tokens, token-to-image cross-attention, an MLP layer, and image-to-token cross-attention that updates the image embedding.
(4) Pixel Decoder: upsamples the output image embedding with a stride of 2. 
Lastly, the model produces matte masks by computing a dot product between the upsampled image embedding and output token embeddings.

\vspace{0.75mm} \noindent \textbf{Motivation.}
While SAM has shown success in segmentation tasks, its pixel decoder, which comprises two straightforward transposed convolutional layers, is prone to generating checkerboard artifacts, especially when handling challenging visual prompts, such as multiple positive and negative points placed near object boundaries or box prompts with imprecise object region delineation, as shown in Figure \ref{fig:cover}.
Furthermore, their upsampled embeddings with a stride of 4 are often insufficient for image matting, which benefits from finer mask feature representations.

\vspace{0.75mm} \noindent \textbf{Hierarchical Pixel Decoder.}
To address these shortcomings, we introduce a hierarchical pixel decoder with a multi-level feature pyramid design, motivated by \cite{(vitmatte)yao2024vitmatte}, as illustrated in Figure \ref{fig:overview}.
The pixel decoder takes an input image and generates multi-resolution feature maps at strides 2, 4, and 8 using a series of simple convolutional layers. The image embedding is sequentially upsampled and concatenated with the corresponding feature maps at each resolution.
The decoder is designed to be highly lightweight, namely adding only 10 ms of computational overhead compared to the original pixel decoder of SAM on a V100 GPU.
Our hierarchical design serves two key purposes:
First, it preserves high-level semantics while refining spatial details, reducing checkerboard artifacts and enhancing robustness to challenging prompts.
Second, it generates high-resolution feature maps with a stride of 2, essential for capturing fine-grained structures in matting.

\vspace{0.75mm} \noindent \textbf{Prompt-Aware Masked Attention.}
To further boost the interactive matting performance, we propose a Prompt-Aware Masked Attention mechanism, inspired by Mask2Former~\cite{(mask2former)cheng2022masked} (See Figure \ref{fig:overview}).
This mechanism allows the model to dynamically focus on the relevant regions within the image based on visual prompts ($e.g.,$ points or boxes), enabling more attention to the areas of interest.

For box prompts, we generate a binary attention mask $\mathcal{M}^{b}$ that indicates the specific bounding box region.
The binary attention mask $\mathcal{M}^{b} \in \{0, -\infty\}$ is defined as:
\begin{equation}
    \mathcal{M}^{b}(x, y) = 
    \begin{cases} 0 & \text{if } (x, y) \in \text{box region} \\ 
    - \infty & \text{otherwise} \end{cases} 
\end{equation}
where \((x, y)\) represents the pixel coordinates.
This forces the model to prioritize the region within the box prompt.

For point prompts, we generate a soft attention mask using a 2D Gaussian map distribution with standard deviation $\sigma$. 
The soft attention mask, $\mathcal{M}^{p} \in [0, 1]$, smoothly weighs the region around the point of interest, ensuring a graded focus that transitions smoothly to the surrounding regions.

The attention mask is incorporated into the cross-attention blocks of the transformer decoder. 
Specifically, the attention mask modulates the attention map as follows:
\begin{equation}
   X_{l} = \begin{cases} \text{softmax}(\mathcal{M}^{b}+Q_{l}K^{\intercal}_{l})V_{l}+X_{l-1} & \text{(box prompt)} \\ 
   \text{softmax}(\mathcal{M}^{p} \odot Q_{l}K^{\intercal}_{l})V_{l}+X_{l-1} & \text{(point prompt)} 
   \end{cases} 
\end{equation}
where $\odot$ denotes element-wise multiplication, $X_{l}$ represents the query feature maps at the $l^{th}$ layer of the decoder, and $Q_{l}$, $K_{l}$, and $V_{l}$ implies the query, key, and value matrices, respectively, at the $l^{th}$ layer.
This mechanism dynamically adjusts the model’s attention according to the visual prompt, leading to performance improvement in prompt-driven interactive scenarios (see Table \ref{tab:analysis_model_attn_dec}).

\vspace{0.75mm} \noindent \textbf{Training.}
We train ZIM using the SA1B-Matte dataset.
From the ground-truth matte label, we extract a box prompt from the given min-max coordinates and randomly sample positive and negative point prompts following \cite{(RITM)sofiiuk2022reviving}.
The model is optimized using the same matte loss functions defined in Eq. (\ref{eq:loss_function}).

\begin{table*}[t]
  \centering
  \begin{adjustbox}{max width=0.92\linewidth}
  \begin{tabular}{c|c|c|c|c|c|c|c|c|c|c|c}
    \toprule
    \multirow{2}{*}{Method} & \multirow{2}{*}{Prompt} & \multicolumn{5}{c|}{\textbf{MicroMat3K Fine-grained}} & \multicolumn{5}{c}{\textbf{MicroMat3K Coarse-grained}} \\
     & & SAD{$\downarrow$} & MSE{$\downarrow$} & MAE{$\downarrow$} & Grad{$\downarrow$} & Conn{$\downarrow$} & SAD{$\downarrow$} & MSE{$\downarrow$} & MAE{$\downarrow$} & Grad{$\downarrow$} & Conn{$\downarrow$} \\
    \midrule
    \multirow{2}{*}{SAM~\cite{(SAM)kirillov2023segment}} & point & 68.076 & 21.651 & 23.307 & 16.496 & 67.730 & 17.093 & 5.569 & 5.756 & 4.800 & 17.035 \\
     & box & 36.086 & 11.057 & 12.714 & 14.867 & 35.834 & 3.516 & 1.044 & 1.231 & 2.551 & 3.450 \\
    \midrule
    \multirow{2}{*}{HQ-SAM~\cite{(HQ-SAM)ke2024segment}} & point & 110.681 & 36.674 & 38.331 & 16.855 & 110.421 & 18.842 & 6.457 & 6.645 & 4.599 & 18.792 \\
     & box & 124.262 & 42.457 & 44.144 & 13.673 & 124.113 & 8.458 & 2.733 & 2.920 & 2.472 & 8.400 \\
    \midrule
    \multirow{2}{*}{Matte-Any~\cite{(matte-any)yao2024matte}} & point & 68.797 & 20.844 & 23.564 & 8.118 & 68.939 & 19.717 & 6.053 & 6.675 & 2.633 & 19.506  \\
    & box & 34.661 & 9.746 & 12.182 & 7.021 & 34.856 & 6.950 & 1.983 & 2.445 & 2.142 & 6.905 \\
    \midrule
    \multirow{2}{*}{Matting-Any~\cite{(matting-any)li2024matting}} & point & 275.398 & 77.335 & 97.141 & 20.019 & 270.722 & 164.145 & 36.187 & 55.943 & 23.244 & 155.780 \\
    & box & 246.214 & 68.372 & 87.617 & 19.185 & 241.597 & 109.639 & 23.780 & 38.662 & 15.841 & 102.439 \\
    \midrule
    \multirow{2}{*}{\textbf{ZIM (ours)}} & point & \textbf{31.286} & \textbf{8.213} & \textbf{10.740} & \textbf{5.324} & \textbf{31.009} & \textbf{6.645} & \textbf{1.788} & \textbf{2.320} & \textbf{1.469} & \textbf{6.472} \\
    & box & \textbf{9.961} & \textbf{1.893} & \textbf{3.426} & \textbf{4.813} & \textbf{9.655} & \textbf{1.860} & \textbf{0.448} & \textbf{0.659} & \textbf{1.281} & \textbf{1.807} \\
    \bottomrule
  \end{tabular}
  \end{adjustbox}
  \caption{
    \textbf{Quantitative comparison} of ZIM and six existing methods on the MicroMat-3K test set, evaluated separately on fine-grained and coarse-grained categories using point and box prompts across five evaluation metrics. 
  }
  \label{tab:comparision_mat}
  \vspace{-1mm}
\end{table*}

\section{MicroMat-3K: Zero-Shot Matting Test Set} \label{sec:4}
We introduce a new test set, named MicroMat-3K, to evaluate zero-shot interactive matting models.
It consists of 3,000 high-resolution images paired with micro-level matte labels.
It includes two types of matte labels: (1) Fine-grained labels ($e.g.,$ hair, tree branches) to primarily evaluate zero-shot matting performance, where capturing intricate details is critical. (2) coarse-grained labels ($e.g.,$ cars, desks) to allow comparison with zero-shot segmentation models, which is still essential in zero-shot matting tasks.
Moreover, It provides pre-defined point prompt sets for positive and negative points and box prompt sets for evaluating interactive scenarios.
More detailed information about the MicroMat-3K is described in the supplementary material.

\section{Experiments}

\subsection{Experimental Setting.}

\vspace{1.0mm} \noindent \textbf{Training Dataset for Label Converter.}
To train the label converter, we collect six publicly available matting datasets ($i.e.,$ AIM-500~\cite{(AIM-500)li2021deep}, AM-2K~\cite{(AM-2K)li2022bridging}, P3M-10K~\cite{(P3M)li2021privacy}, RWP-636~\cite{(MGM)(RWP-636)yu2021mask}, HIM-2K~\cite{(HIM-2K)sun2022human}, and RefMatte~\cite{(RefMatte)li2023referring}), consisting of 20,591 natural images and 118,749 synthetic images in total.
For non-transformable samples, we extract coarse object categories ($e.g.,$ car and desk) from the ADE20K dataset~\cite{(ade20k)zhou2017scene}, sampling 187,063 masks from 17,768 images.

\vspace{0.75mm} \noindent \textbf{Evaluation Metrics.}
We use widely adopted evaluation metrics for the image matting task, including Sum of Absolute Difference (SAD), Mean Squared Error (MSE), Gradient Error (Grad), and Connectivity Error (Conn).

\begin{figure*}[t]
    \centering
    \includegraphics[width=0.94\linewidth]{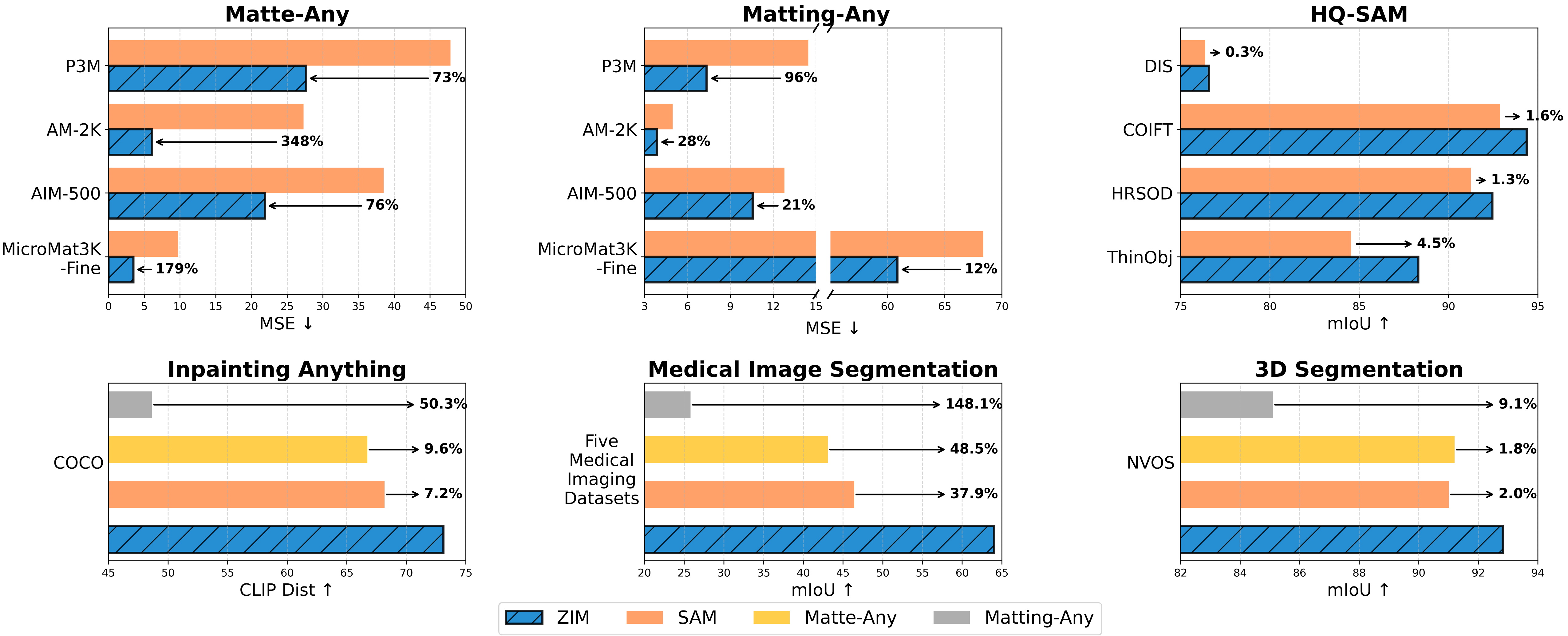}
    \caption{
        \textbf{Downstream Transferability Evaluation} by replacing SAM with ZIM as the segmentation foundation model in various downstream tasks:
        Matte-Any~\cite{(matte-any)yao2024matte} and Matting-Any~\cite{(matting-any)li2024matting} are evaluated on MicroMat3K-Fine, AIM-500~\cite{(AIM-500)li2021deep}, AM-2K~\cite{(AM-2K)li2022bridging}, and P3M-500-NP~\cite{(P3M)li2021privacy} using the MSE metric. 
        HQ-SAM~\cite{(HQ-SAM)ke2024segment} is assessed on DIS~\cite{(DIS)qin2022highly}, COIFT~\cite{(COIFT)(ThinObject)liew2021deep}, HRSOD~\cite{(HRSOD)zeng2019towards}, and ThinObject~\cite{(COIFT)(ThinObject)liew2021deep} using mIoU. 
        Inpainting Anything~\cite{(inpaint_anything)yu2023inpaint} is tested on COCO~\cite{(coco)lin2014microsoft} using CLIP distance.  
        Following Medical Image Segmentation evaluation protocol~\cite{(medi_sam)mazurowski2023segment} with five different prompting modes (mIoU).  
        3D segmentation follows the SA3D~\cite{(SA3D)cen2023segment} framework and is evaluated on NVOS~\cite{(NVOS)ren-cvpr2022-nvos}.  
    }
    \label{fig:transfer_sam_zim}
    \vspace{-1mm}
\end{figure*}

\vspace{0.75mm} \noindent \textbf{Implementation Details for Label Converter.}
The label converter model is based on MGMatting~\cite{(MGM)(RWP-636)yu2021mask} with Hiera-base-plus~\cite{(hiera)ryali2023hiera} backbone network.
For training the converter, we set the input size to 1024$\times$1024, a batch size of 16, and a learning rate of 0.001 with cosine decay scheduling using the AdamW optimizer~\cite{(adamw)loshchilov2017decoupled}.
The training process runs for 500K iterations with the probability parameter $p$ of 0.5 and the loss weight $\lambda$ of 10.

\vspace{0.75mm} \noindent \textbf{Implementation Details for ZIM.}
For the ZIM model, we use the same image encoder ($i.e.,$ ViT-B~\cite{(vit)dosovitskiy2020image}) and prompt encoder as SAM.
Leveraging the pre-trained weights from SAM, we fine-tune the ZIM model on 1\% of the SA1B-Matte dataset, which amounts to approximately 2.2M matte labels.
We set the input size to 1024$\times$1024, batch size to 16, a learning rate to 0.00001 with cosine decay scheduling using the AdamW optimizer~\cite{(adamw)loshchilov2017decoupled}, and training iterations to 500K.
The loss weight $\lambda$ is set to 10 and the $\sigma$ for the point-based attention mask is set to 21 by default.

\subsection{Experimental Results}
We evaluate ZIM against four related methods on MicroMat3K: SAM~\cite{(SAM)kirillov2023segment}, HQ-SAM~\cite{(HQ-SAM)ke2024segment}, Matte-Any~\cite{(matte-any)yao2024matte}, and Matting-Any~\cite{(matting-any)li2024matting}.
All methods use the ViT-B~\cite{(vit)dosovitskiy2020image} backbone network.
Table \ref{tab:comparision_mat} presents the evaluation scores across five metrics for both point and box prompts on fine-grained and coarse-grained masks.
For coarse-grained results, SAM achieves reasonable zero-shot performance, while other methods ($e.g.,$ Matting-Any and HQ-SAM) struggle to generalize to unseen objects. This is likely due to their fine-tuning on macro-level labeled datasets, which degrades their zero-shot capabilities.
The qualitative results in Figure \ref{fig:cover} show that these methods often produce macro-level outputs, even provided with micro-level prompts.
Moreover, SAM tends to suffer from checkerboard artifacts when challenging prompts are introduced. In contrast, ZIM generates a more robust quality of masks, due to our hierarchical feature pyramid decoder. 
The fine-grained results in Table \ref{tab:comparision_mat} highlight ZIM's superiority in producing high-quality matting outputs while maintaining strong zero-shot capabilities.

\subsection{Downstream Transferability}
We assess the transferability of ZIM compared to SAM across various downstream tasks. 
Specifically, we first integrate ZIM into existing matting methods, Matte-Any~\cite{(matte-any)yao2024matte} and Matting-Any~\cite{(matting-any)li2024matting}, and observe significantly improved box prompting MSE results across diverse matting benchmarks, including MicroMat3K-Fine, AIM-500~\cite{(AIM-500)li2021deep}, AM-2K~\cite{(AM-2K)li2022bridging}, and P3M-500-NP~\cite{(P3M)li2021privacy}. 
Likewise, replacing SAM with ZIM in HQ-SAM~\cite{(HQ-SAM)ke2024segment} notably enhances performance in fine-grained segmentation datasets, including DIS~\cite{(DIS)qin2022highly}, COIFT~\cite{(COIFT)(ThinObject)liew2021deep}, HRSOD~\cite{(HRSOD)zeng2019towards}, and ThinObject~\cite{(COIFT)(ThinObject)liew2021deep}.

Moreover, we extend our analysis to broader vision tasks, including image inpainting, medical image segmentation, and 3D segmentation. 
Specifically, we evaluate image inpainting using the Inpainting Anything framework~\cite{(inpaint_anything)yu2023inpaint} on the COCO dataset~\cite{(coco)lin2014microsoft} via the CLIP Distance metric~\cite{ekin2024clipaway, yildirim2023instinpaint}.
For medical image segmentation, we utilize the zero-shot evaluation protocol from~\cite{(medi_sam)mazurowski2023segment} on five diverse medical imaging datasets~\cite{Antonelli_2022,SONG2022106706,ultrasound-nerve-segmentation,x-ray}, measured by mean IoU (mIoU).
Additionally, we assess 3D segmentation performance within the SA3D framework~\cite{(SA3D)cen2023segment} on the NVOS dataset~\cite{(NVOS)ren-cvpr2022-nvos}. 
These tasks inherently require highly precise segmentation masks for optimal outcomes. 
While replacing SAM with existing matting models ($e.g.,$ Matte-Any and Matting-Any) results in significant performance drops due to limited generalization capabilities, employing ZIM consistently boosts zero-shot performance across these diverse scenarios, highlighting its robust generalization and precise mask representation capabilities. Comprehensive qualitative and quantitative results for these tasks are provided in the supplementary materials.

\begin{figure}[t]
    \centering
    \begin{subfigure}[b]{0.99\linewidth}
        \centering
        \includegraphics[width=\linewidth]{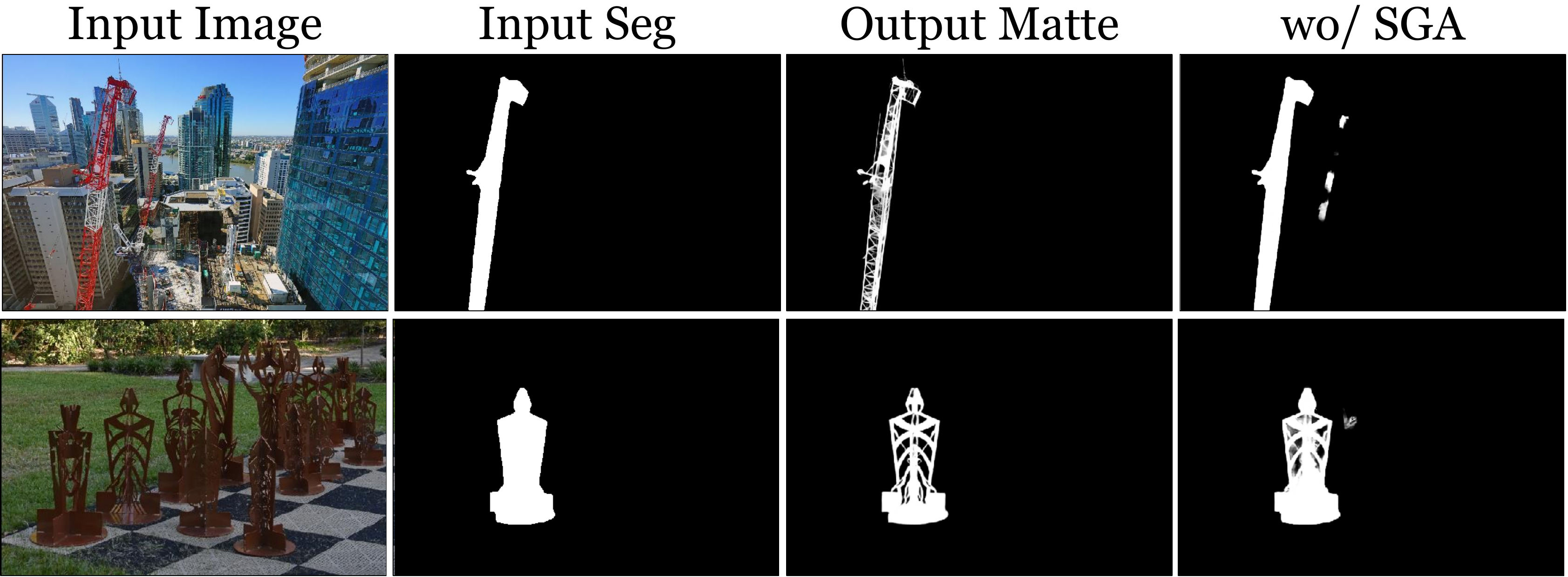} 
        \caption{}
        \label{fig:analysis_SGA}  
        \vspace{1mm}
    \end{subfigure}
    \begin{subfigure}[b]{0.99\linewidth}
        \centering
        \includegraphics[width=\linewidth]{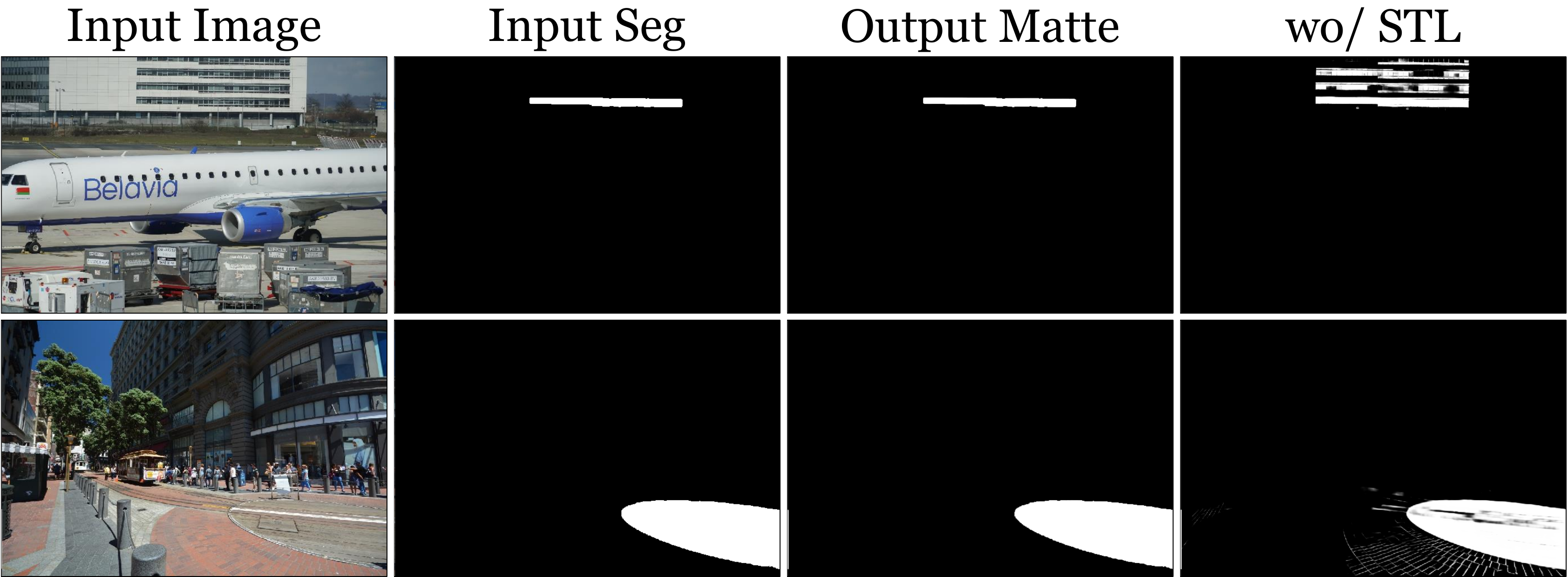} 
        \caption{}
        \label{fig:analysis_STL} 
    \end{subfigure}
    \vspace{-2mm}
    \caption{
    \textbf{Qualitative analysis of key components of the label converter}: (a) without Spatial Generalization Augmentation (SGA) and (b) without Selective Transformation Learning (STL).
    }
    \label{fig:analysis_converter}
    \vspace{-2mm}
\end{figure}

\begin{table}[t]
  \centering
  \begin{adjustbox}{max width=\linewidth}
  \begin{tabular}{c|c|c|c|c|c|c|c}
    \toprule
    \multirow{2}{*}{SGA} & \multirow{2}{*}{STL} & \multicolumn{3}{c|}{\textbf{Fine-grained}} & \multicolumn{3}{c}{\textbf{Coarse-grained}} \\
     & & SAD{$\downarrow$} & MSE{$\downarrow$} & Grad{$\downarrow$} & SAD{$\downarrow$} & MSE{$\downarrow$} & Grad{$\downarrow$} \\
    \midrule
    &  & 3.324 & 0.276 & 2.664 & 0.716 & 0.117 & 0.684 \\
    \checkmark &  & 2.440 & 0.122 & 2.139 & 0.697 & 0.092 & 0.634 \\
    & \checkmark &  3.153 & 0.239 & 2.457 & 0.635 & 0.089 & 0.653 \\
    \checkmark & \checkmark & \textbf{1.999} & \textbf{0.080} & \textbf{1.771} & \textbf{0.281} & \textbf{0.021} & \textbf{0.399} \\
    \bottomrule
  \end{tabular}
  \end{adjustbox}
  \caption{
    \textbf{Quantitative analysis of key components of the label converter}: Spatial Generalization Augmentation (SGA) and Selective Transformation Learning (STL).
  }
  \label{tab:analysis_converter}
  \vspace{-2mm}
\end{table}

\section{Ablation Study}
In this section, we analyze the impact of the key components of our method using the MicroMat-3K test set.

\vspace{2.0mm} \noindent \textbf{Analysis of Label Converter.}
To analyze the effect of Spatial Generalization Augmentation (SGA) and Selective Transformation Learning (STL) strategies, we conduct an ablation study by removing each component individually.
The SGA is designed to enhance the generalization ability of the converter by simulating diverse input patterns, particularly beneficial given that the converter is trained on macro-level labeled datasets.
Without the SGA, the converter struggles to produce clear matte labels for unseen objects, as shown in Figure \ref{fig:analysis_SGA}.
In addition, the STL is designed to help the converter avoid unnecessary label conversion for coarse objects.
Without the STL, the converter attempts to transform every segmentation label into a matte label, resulting in noisy outputs for unseen coarse objects, as shown in Figure \ref{fig:analysis_STL}.
The quantitative results in Table \ref{tab:analysis_converter} confirm that using both strategies yields the best label conversion performance on the MicroMat-3K test set.

\vspace{1.2mm} \noindent \textbf{Analysis of ZIM Model.}
We conduct experiments to analyze the effect of the prompt-aware masked attention and hierarchical mask decoder.
The prompt-aware masked attention is designed to direct the model's focus on the regions of interest to improve the promptable matting performance.
Table \ref{tab:analysis_model_attn_dec} shows that leveraging the masked attention yields a substantial improvement to our ZIM model.
In addition, the hierarchical mask decoder is designed to produce more robust and higher-resolution mask feature maps to alleviate checkerboard artifacts and capture finer representation, simultaneously.
Its effectiveness is particularly evident in reducing the gradient error for fine-grained objects in Table \ref{tab:analysis_model_attn_dec}, since the enhanced pixel decoder generates more solid and detailed mask outputs. 
Notably, the decoder remains lightweight, adding only 10 ms of additional inference time.

\begin{table}[t]
    \centering
    \begin{subtable}{\linewidth}
        \centering
        \begin{adjustbox}{max width=\linewidth}
          \begin{tabular}{c|c|c|c|c|c|c|c}
            \toprule
            \multirow{2}{*}{Attn} & \multirow{2}{*}{Dec} & \multicolumn{3}{c|}{\textbf{Fine-grained}} & \multicolumn{3}{c}{\textbf{Coarse-grained}} \\
             & & SAD{$\downarrow$} & MSE{$\downarrow$} & Grad{$\downarrow$} & SAD{$\downarrow$} & MSE{$\downarrow$} & Grad{$\downarrow$} \\
            \midrule
            &  & 13.623 & 2.718 & 6.516 & 2.071 & 0.474 & 1.526 \\
            \checkmark &  & 13.198 & 2.504 & 6.445 & 2.049 & 0.471 & 1.486 \\
             & \checkmark & 11.074 & 2.094 & 5.401 & 2.069 & 0.487 & 1.355 \\
            \checkmark & \checkmark & \textbf{9.961} & \textbf{1.893} & \textbf{4.813} & \textbf{1.860} & \textbf{0.448} & \textbf{1.281} \\
            \bottomrule
          \end{tabular}
          \end{adjustbox}
      \caption{}
      \label{tab:analysis_model_attn_dec}
    \end{subtable}%
    \vspace{1mm}
    \begin{subtable}{\linewidth}
        \centering
        \begin{adjustbox}{max width=\linewidth}
          \begin{tabular}{c|c|c|c|c|c|c|c}
            \toprule
            \multicolumn{2}{c|}{\textbf{Attn Mask}} & \multicolumn{3}{c|}{\textbf{Fine-grained}} & \multicolumn{3}{c}{\textbf{Coarse-grained}} \\
            T2I & I2T & SAD{$\downarrow$} & MSE{$\downarrow$} & Grad{$\downarrow$} & SAD{$\downarrow$} & MSE{$\downarrow$} & Grad{$\downarrow$} \\
            \midrule
            &  & 11.074 & 2.094 & 5.401 & 2.069 & 0.487 & 1.355 \\
            \checkmark &  & \textbf{9.961} & \textbf{1.893} & \textbf{4.813} & \textbf{1.860} & \textbf{0.448} & \textbf{1.281} \\
             & \checkmark & 12.526 & 2.658 & 6.032 & 2.353 & 0.554 & 1.481 \\
            \checkmark & \checkmark & 10.437 & 1.997 & 5.066 & 1.999 & 0.470 & 1.306 \\
            \bottomrule
          \end{tabular}
          \end{adjustbox}
      \caption{}
      \label{tab:analysis_model_attn_mask}
    \end{subtable}
    \vspace{3mm}
        \begin{subtable}{\linewidth}
        \centering
        \begin{adjustbox}{max width=\linewidth}
          \begin{tabular}{c|c|c|c|c|c|c|c}
            \toprule
            \multirow{2}{*}{Model} & \multirow{2}{*}{Trainset} & \multicolumn{3}{c|}{\textbf{Fine-grained}} & \multicolumn{3}{c}{\textbf{Coarse-grained}} \\
             & & SAD{$\downarrow$} & MSE{$\downarrow$} & Grad{$\downarrow$} & SAD{$\downarrow$} & MSE{$\downarrow$} & Grad{$\downarrow$} \\
            \midrule
            \multirow{2}{*}{ZIM} & SA1B-Matte & \textbf{9.961} & \textbf{1.893} & \textbf{4.813} & \textbf{1.860} & \textbf{0.448} & \textbf{1.281} \\
             & Public-Matte & 120.571 & 38.332 & 6.730 & 9.506 & 2.760 & 1.688 \\
             \midrule
             Matting- & SA1B-Matte & 41.242 & 12.267 & 7.707 & 4.626 & 1.284 & 1.552 \\
             Any~\cite{(matting-any)li2024matting} & Public-Matte & 246.214 & 68.372 & 19.185 & 109.639 & 23.780 & 15.841 \\
            \bottomrule
          \end{tabular}
          \end{adjustbox}
      \caption{}
      \label{tab:analysis_trainset}
    \end{subtable}%
    \label{tab:analysis_model}
    \vspace{-8mm}
    \caption{\textbf{Analysis of ZIM} using box prompt evaluations: 
    (a) Effect of \texttt{Attn} (prompt-aware masked attention) and \texttt{Dec} (hierarchical pixel decoder).
    (b) Effect of the masked attention in \texttt{T2I} (token to image) and \texttt{I2T} (image to token) cross-attention layers.
    (c) Effect of trainset: our SA1B-Matte and public matting datasets.
    }
    \vspace{-1mm}
\end{table}

Moreover, we delve into the effect of prompt-aware masked attention in our transformer decoder, which comprises two kinds of cross-attention layers (see Figure \ref{fig:overview}): token-to-image (t2i) updating token embeddings (as queries) and image-to-token (i2t) updating the image embedding (as queries). 
As a result in Table \ref{tab:analysis_model_attn_mask}, applying masked attention to only the t2i layer leads to a meaningful improvement.
This suggests that focusing attention on tokens based on visual prompts in the t2i layer enhances their ability to capture relevant features. 
In contrast, applying attention to specific regions within the image embedding in the i2t layer may disturb the capture of global features.

\vspace{1.2mm} \noindent \textbf{Analysis of Training Dataset.}
Table \ref{tab:analysis_trainset} investigates the influence of the training dataset on ZIM's zero-shot matting performance. When ZIM is trained on publicly available matting datasets, which predominantly contain macro-level masks, the performance on the micro-level MicroMat3K dataset significantly degrades. 
This result underscores the necessity of training with micro-level mask annotations for effective generalization to multi-granularity details. Despite Matting-Any~\cite{(matting-any)li2024matting} also being trained on SA1B-Matte, ZIM still surpasses its performance by a considerable margin, highlighting the advancements of our network architectural improvements for the zero-shot interactive matting task.

\begin{figure}[t]
    \centering
    \includegraphics[width=\linewidth]{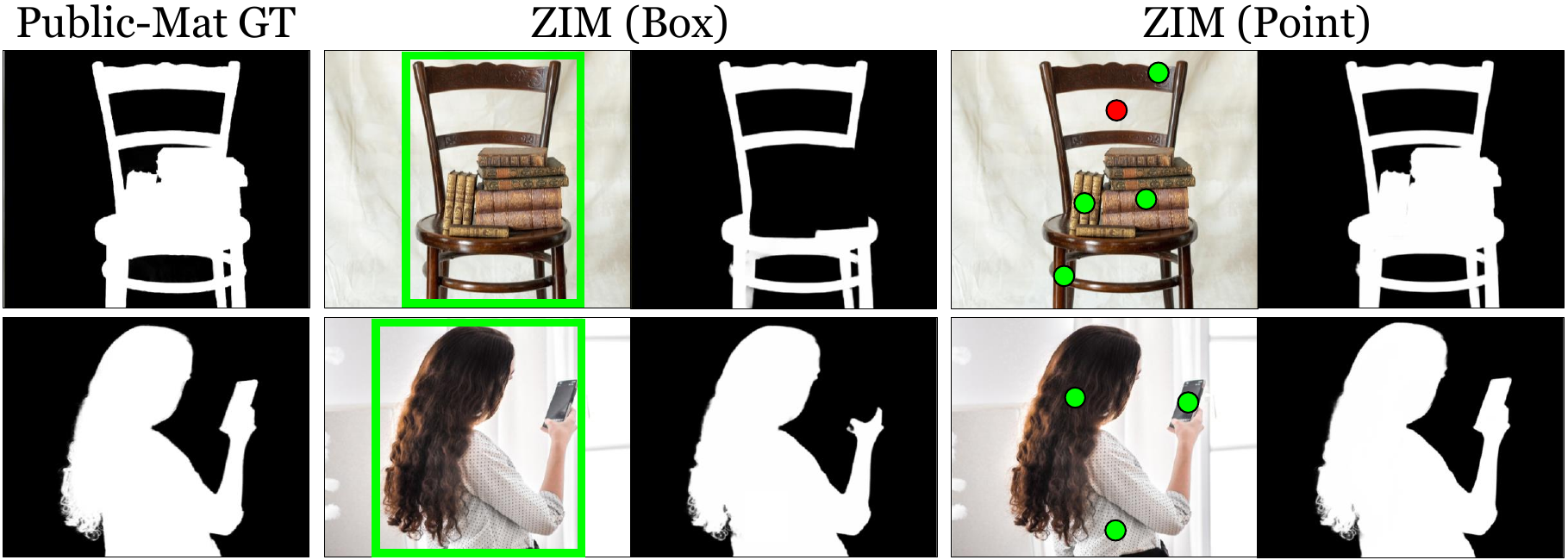}
    \caption{
        \textbf{Domain shift} between traditional public matting testsets and ZIM trained with SA1B-Matte (object-level).
    } 
    \label{fig:domain_gap_gt_zim}
    \vspace{-1mm}
\end{figure}

\begin{figure}[t]
    \centering
    \includegraphics[width=\linewidth]{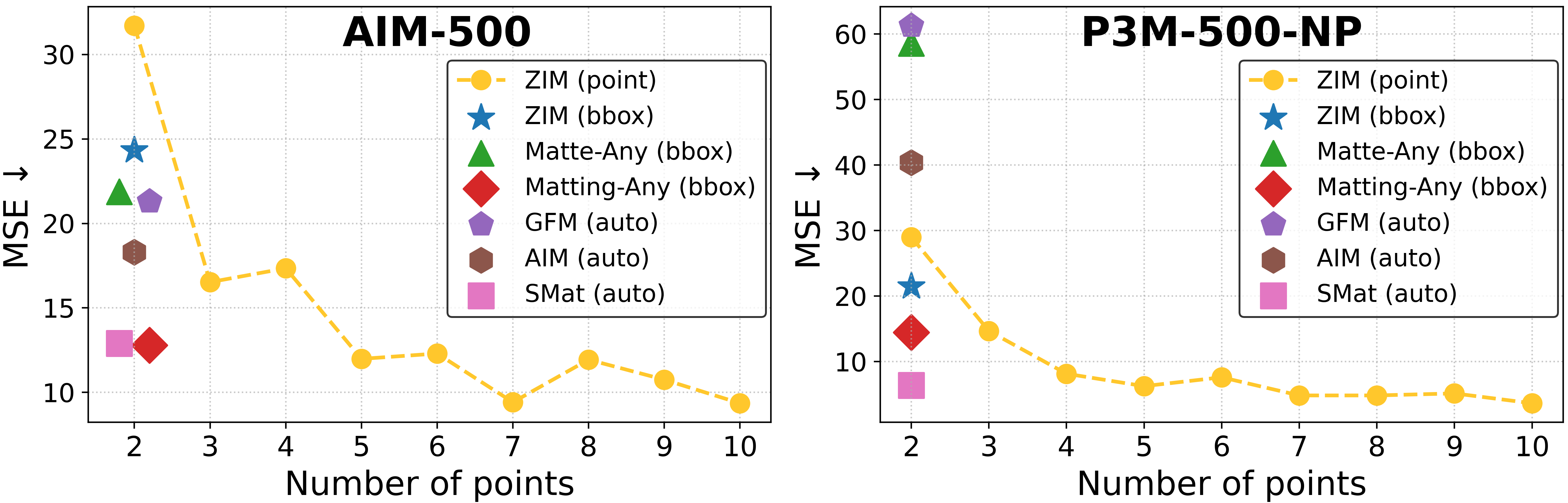}
    \caption{
        \textbf{Quantitative comparison} of ZIM (with varying numbers of point prompts) and existing matting methods (Matte-Any~\cite{(matte-any)yao2024matte}, Matting-Any~\cite{(matting-any)li2024matting}, GFM~\cite{(GFM)li2022bridging}, AIM~\cite{(AIM-500)li2021deep}, and SMat~\cite{(smat)ye2024unifying}) on the traditional matting test sets (AIM-500~\cite{(AIM-500)li2021deep} and P3M-500-NP~\cite{(P3M)li2021privacy}). The ``auto" implies prompt-free mode.
    } 
    \label{fig:zim_noc}
    \vspace{-1mm}
\end{figure}

\vspace{1.2mm} \noindent \textbf{Discussion on Domain Shift.}
There is a distinct domain shift between traditional matting test sets and ZIM trained with SA1B-Matte.
Traditional matting test sets~\cite{(AIM-500)li2021deep,(P3M)li2021privacy} typically regard entire salient objects as foreground, whereas ZIM mainly focuses on object- and part-level matting (Figure \ref{fig:domain_gap_gt_zim}). 
This mismatch leads to substantial performance penalties for ZIM under box prompting on these test sets (Figure \ref{fig:domain_gap_gt_zim}), which is inherent from SAM's prompt ambiguity issue.
However, by leveraging dense multiple-point prompting, ZIM effectively mitigates this discrepancy, surpassing existing methods~\cite{(matte-any)yao2024matte,(matting-any)li2024matting,(GFM)li2022bridging,(AIM-500)li2021deep,(smat)ye2024unifying}, even some of which are explicitly trained on datasets similar to the evaluation domain (Figure \ref{fig:zim_noc}). 
This highlights the adaptability of our zero-shot interactive matting modeling.

\section{Conclusion, Limitation, and Future Work}

In this paper, we presented a pioneering zero-shot image matting model that advances the field by generating precise, fine-grained matte masks. 
We addressed the limitations of SAM, which struggles with high-detail segmentation tasks, by introducing a novel label conversion method and enhancing the network architecture with a hierarchical pixel decoder and prompt-aware masked attention mechanism.
However, ZIM inherits inherent shortcomings from SAM, including ambiguous visual prompt handling and robustness in uncertain predictions. Future research should aim at resolving these distinct challenges through innovative approaches to prompt design and uncertainty-based modeling.
We hope that the research community will continue to build on this work, exploring new applications in computer vision and enhancing its zero-shot matting performance.

\nocite{kim2024eclipse}
\nocite{(grounded_sam)ren2024grounded}
\nocite{(medi_sam)mazurowski2023segment}
\nocite{(maskformer)cheng2021per}
\nocite{(indexmat)lu2019indices}
\nocite{li2020natural}
\nocite{park2022matteformer}
\nocite{wei2021improved}
\nocite{ding2022deep}
\nocite{bai2007geodesic}
\nocite{kim2024towards}
\nocite{grady2005random}
\nocite{(RefMatte)li2023referring}
\nocite{kim2023devil}
\nocite{liu2021tripartite}
\nocite{wang2020solov2}
\nocite{yang2022unified}
\nocite{chen2018semantic}
\nocite{sun2021semantic}

{
    \small
    \bibliographystyle{ieeenat_fullname}
    \bibliography{ms}
}

\clearpage

\twocolumn[{%
\renewcommand\twocolumn[1][]{#1}%
\maketitle
\begin{center}
    \includegraphics[width=0.97\linewidth]{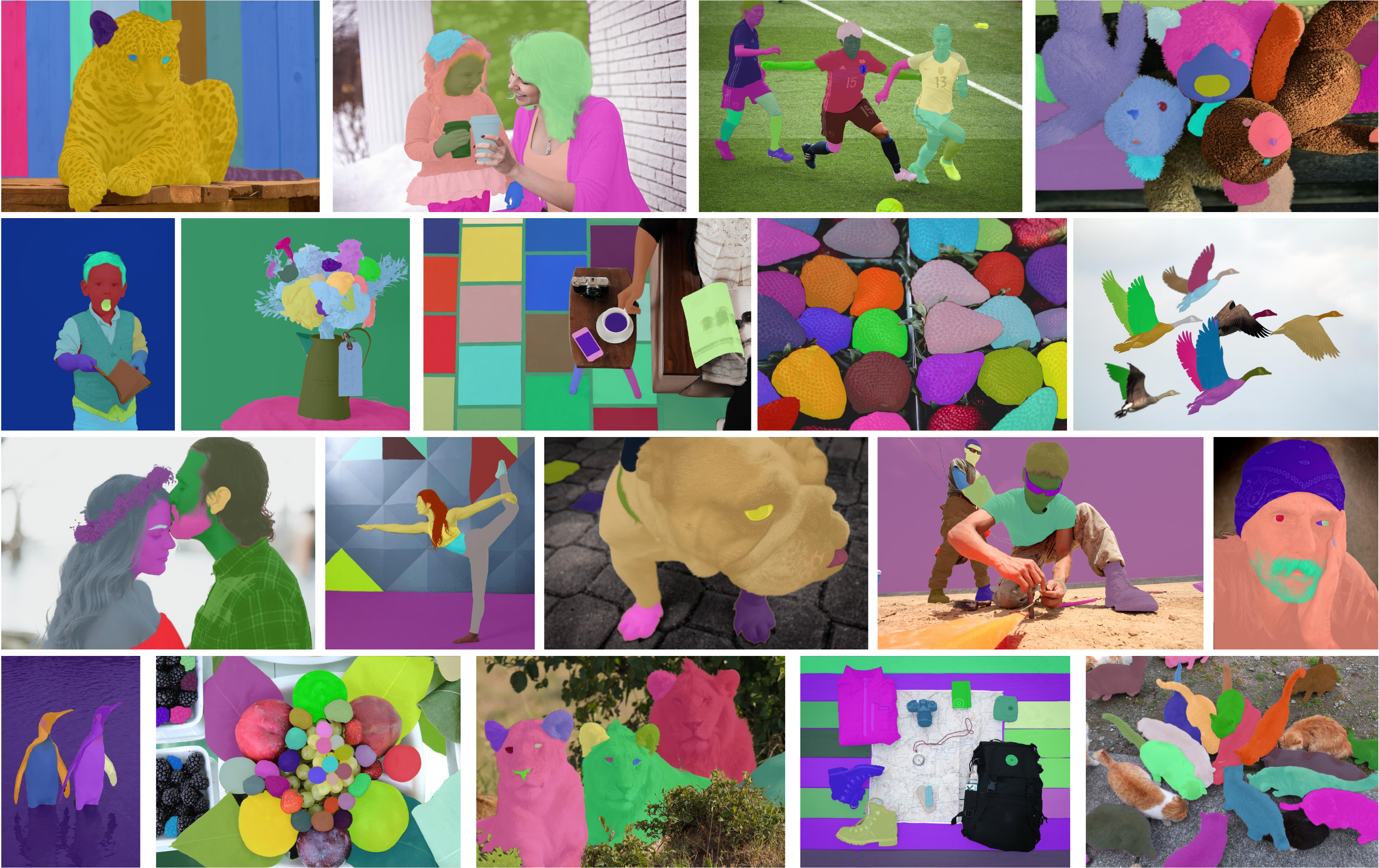}
    \captionof{figure}{
        \textbf{Visualization of samples} from the MicroMat-3K test set, providing a high-quality benchmark for zero-shot matting models.
    }
    \label{fig:micromat}
    \vspace{3mm}
\end{center}
}]

\appendix
\renewcommand{\thesection}{\Alph{section}}
\section*{Appendix}

\section{MicroMat-3K Details}
Inspired by the data engine procedure in the SAM~\cite{(SAM)kirillov2023segment}, we constructed the MicroMat-3K test set.
The dataset construction involved four key steps:
(1) We collected high-resolution images from the DIV2K dataset~\cite{(DIV2K)agustsson2017ntire}, which is originally intended for super-resolution tasks.
(2) We generated pseudo-segmentation labels using automatic mask generation of SAM, powered by a large backbone model to ensure high-quality segmentation.
(3) We transformed these segmentation labels into matte labels using our label converter, which is also powered by a large backbone network. This step provided an initial set of pseudo-matte labels for each image.
(4) Our annotators inspected the pseudo-matte labels. High-quality labels were directly retained as ground-truth annotations, while any low-quality matte labels were manually revised to ensure high-fidelity ground-truth labels.
Additionally, we categorized the final matte labels into two classes: fine-grained masks (750 samples) and coarse-grained masks (2250 samples). 
Figure \ref{fig:micromat} showcases visualization examples from MicroMat-3K, which provides diverse and high-quality micro-level matte labels.

\section{Downstream Task}

\subsection{Zero-Shot Image Matting}
\begin{figure*}[t]
    \centering
    \includegraphics[width=\linewidth]{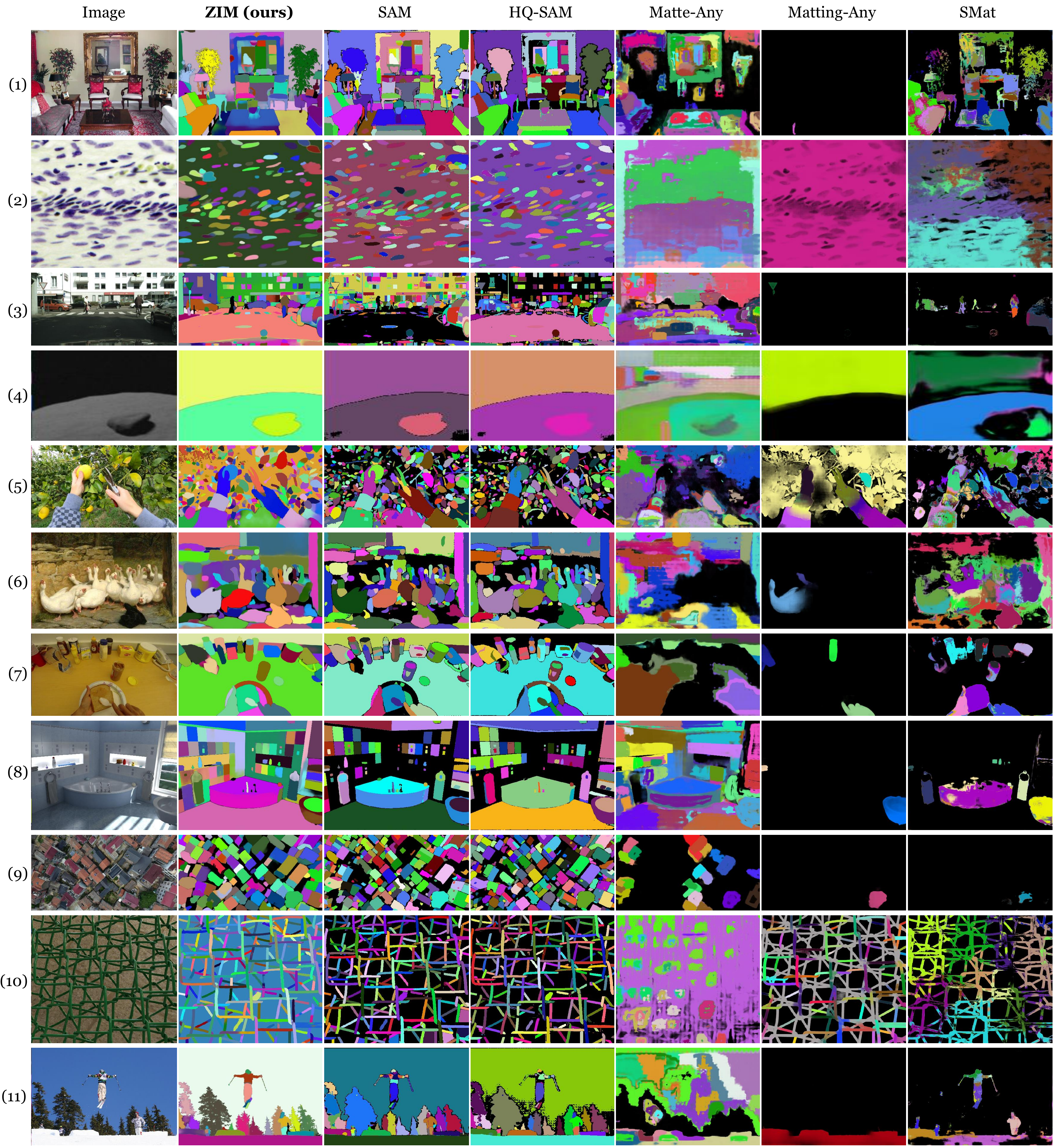}
    \caption{
    \textbf{Qualitative samples of automatic mask generation results} on (1) ADE20K~\cite{(ade20k)zhou2017scene}, (2) BBBC038v1~\cite{(BBBC038v1)caicedo2019nucleus}, (3) Cityscapes~\cite{(Cityscapes)cordts2016cityscapes}, (4) DOORS~\cite{(DOORS)pugliatti2022doors}, (5) EgoHOS~\cite{(EgoHOS)zhang2022fine}, (6) DRAM~\cite{(DRAM)cohen2022semantic}, (7) GTEA~\cite{(GTEA)li2015delving,(GTEA)fathi2011learning}, (8) Hypersim~\cite{(Hypersim)roberts2021hypersim}, (9) IBD~\cite{(IBD)chen20223}, (10) iShape~\cite{(iShape)yang2021ishape}, and (11) COCO~\cite{(coco)lin2014microsoft} datasets.} 
    \label{fig:amg_1}
    \vspace{10mm}
\end{figure*}

\begin{figure*}[t]
    \centering
    \includegraphics[width=0.95\linewidth]{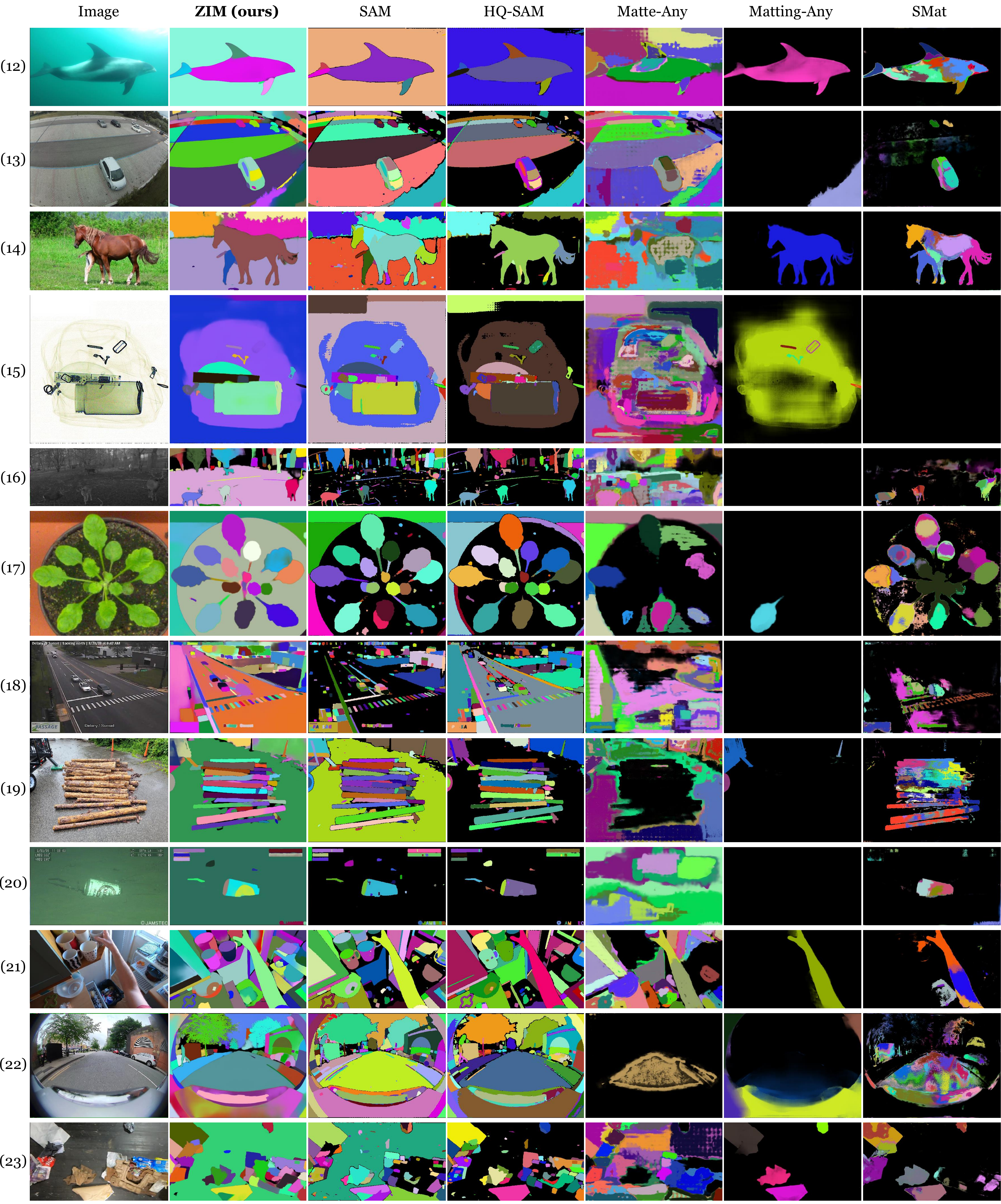}
    \caption{
    (continue) \textbf{Qualitative samples of automatic mask generation results} on (12) NDD20~\cite{(ndd20)trotter2020ndd20}, (13) NDISPark~\cite{(NDISPark)ciampi_ndispark_6560823,(NDISPark)ciampi2021domain}, (14) OVIS~\cite{(OVIS)qi2022occluded}, (15) PIDRay~\cite{(PIDRay)wang2021towards}, (16) Plittersdorf~\cite{(Plittersdorf)haucke2022socrates}, (17) PPDLS~\cite{(PPDLS)minervini2016finely}, (18) STREETS~\cite{(STREETS)snyder2019streets}, (19) TimberSeg~\cite{(TimberSeg)fortin2022instance}, (20) TrashCan~\cite{(TrashCan)hong2020trashcan}, (21) VISOR~\cite{(VISOR)darkhalil2022epic,(VISOR)damen2022rescaling}, (22) WoodScape~\cite{(WoodScape)yogamani2019woodscape}, and (23) ZeroWaste-f~\cite{(ZeroWaste)bashkirova2022zerowaste} datasets.}  
    \label{fig:amg_2}
\end{figure*}

To showcase the generalizability of ZIM, we evaluate it across 23 diverse datasets, including ADE20K~\cite{(ade20k)zhou2017scene}, BBBC038v1~\cite{(BBBC038v1)caicedo2019nucleus}, Cityspcaes~\cite{(Cityscapes)cordts2016cityscapes}, DOORS~\cite{(DOORS)pugliatti2022doors}, EgoHOS~\cite{(EgoHOS)zhang2022fine}, DRAM~\cite{(DRAM)cohen2022semantic}, GTEA~\cite{(GTEA)li2015delving,(GTEA)fathi2011learning}, Hypersim~\cite{(Hypersim)roberts2021hypersim}, IBD~\cite{(IBD)chen20223}, iShape~\cite{(iShape)yang2021ishape}, COCO~\cite{(coco)lin2014microsoft}, NDD20~\cite{(ndd20)trotter2020ndd20}, NDISPark~\cite{(NDISPark)ciampi_ndispark_6560823,(NDISPark)ciampi2021domain}, OVIS~\cite{(OVIS)qi2022occluded}, PIDRay~\cite{(PIDRay)wang2021towards}, Plittersdorf~\cite{(Plittersdorf)haucke2022socrates}, PPDLS~\cite{(PPDLS)minervini2016finely}, STREETS~\cite{(STREETS)snyder2019streets}, TimberSeg~\cite{(TimberSeg)fortin2022instance}, TrashCan~\cite{(TrashCan)hong2020trashcan}, VISOR~\cite{(VISOR)darkhalil2022epic,(VISOR)damen2022rescaling}, WoodScape~\cite{(WoodScape)yogamani2019woodscape}, and Zero Waste-f~\cite{(ZeroWaste)bashkirova2022zerowaste}.
Using the \texttt{Automatic Mask Generation} strategy introduced by SAM~\cite{(SAM)kirillov2023segment}, we apply a regular grid of point prompts to each image and perform post-processing with thresholding and non-maximum suppression (NMS) to generate the final matting masks.
Figure \ref{fig:amg_1} and Figure \ref{fig:amg_2} show that ZIM produces high-quality \emph{matte anything} results for all datasets with considerably detailed matte quality and powerful generalization capability.
Although SAM shows powerful generalization capability, the output mask is the coarse quality.
In addition, existing interactive matting methods ($i.e.,$ Matte-Any~\cite{(matte-any)yao2024matte}, Matting-Any~\cite{(matting-any)li2024matting}, and SMat~\cite{(smat)ye2024unifying}) often fail to generalize the unseen data.

\begin{figure*}[t]
    \centering
    \includegraphics[width=\linewidth]{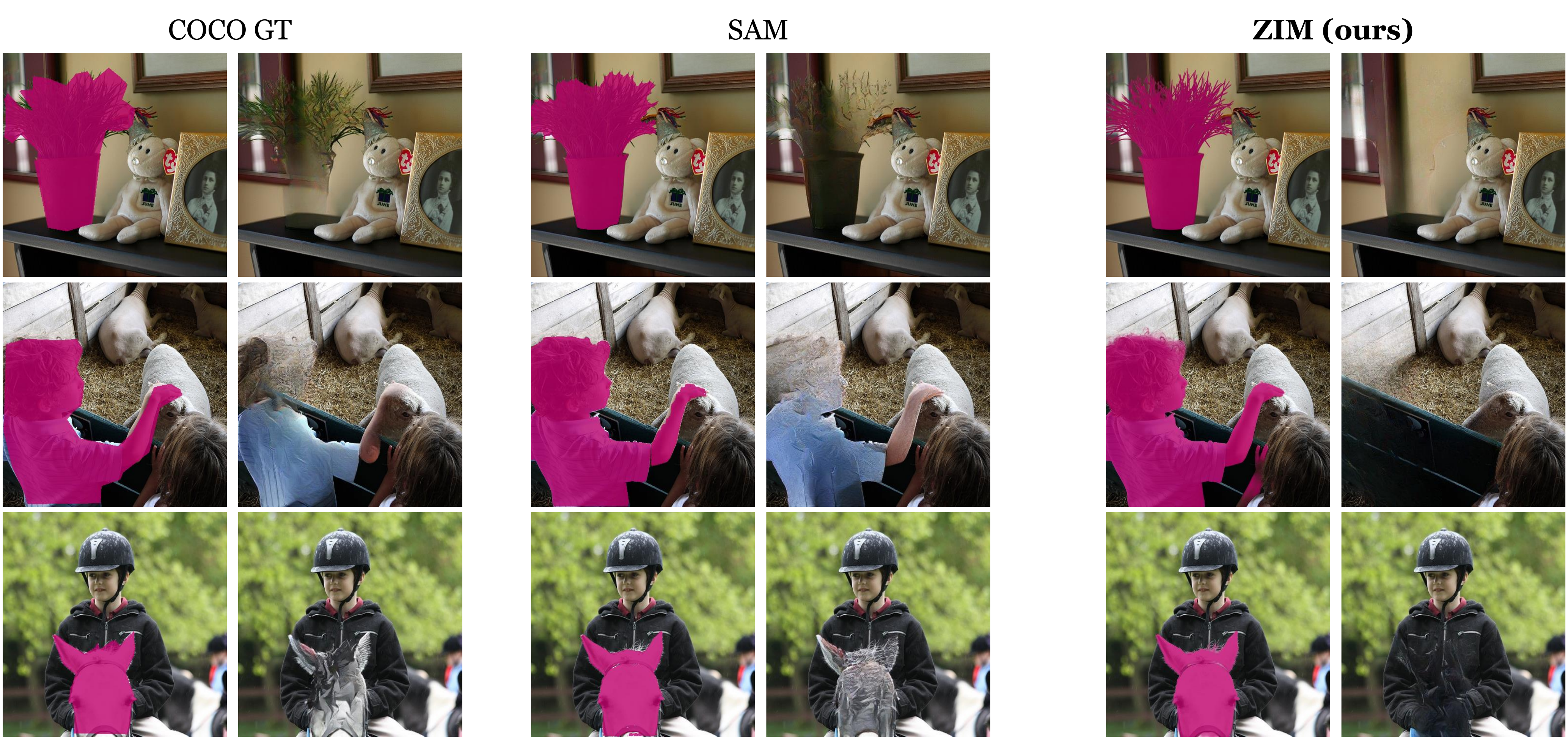}
    \caption{
        \textbf{Qualitative results} of three kinds of input masks ($i.e.,$ COCO ground-truth~\cite{(coco)lin2014microsoft}, SAM~\cite{(SAM)kirillov2023segment}, and ZIM) along with their corresponding image inpainting results using the Inpainting Anything framework~\cite{(inpaint_anything)yu2023inpaint}.    
        }
    \label{fig:inpainting}
\end{figure*}

\subsection{Image Inpainting}
Image inpainting is an important application in generative AI, where precise mask generation plays a critical role in removing or reconstructing parts of an image. 
Following the Inpaint Anything framework~\cite{(inpaint_anything)yu2023inpaint}, we guide SAM and ZIM masks to the inpainting model~\cite{(LaMa)suvorov2022resolution}.
As shown in Figure \ref{fig:inpainting}, ZIM produces more accurate object masks than SAM, leading to significantly better inpainting results.
For complex objects like flowerpots and hair, SAM’s coarse masks fail to remove the object cleanly, leaving noticeable artifacts. In contrast, our precise matting masks enable the inpainting model to smoothly remove objects without artifacts.
To provide quantitative analysis, we use CLIP Distance and CLIP Accuracy~\cite{ekin2024clipaway, yildirim2023instinpaint} as evaluation metrics. CLIP Distance measures the similarity between the source and inpainted regions, where a larger distance indicates better removal. CLIP Accuracy evaluates the change in class predictions after removal, considering the task successful if the original class is absent from Top-1\//3\//5 predictions. Following \cite{yildirim2023instinpaint}, we use the text prompt \texttt{a photo of a \{category name\}}.
Table \ref{tab:downstream_inpaint} shows that ZIM outperforms SAM on both metrics by better preserving surrounding context after inpainting through enhanced mask quality.
\begin{table}[t]
  \centering
  \begin{adjustbox}{max width=\linewidth}
  \begin{tabular}{c|c|ccc}
    \toprule
    \multirow{2}{*}{Mask} & \multirow{2}{*}{\textbf{CLIP Dist $\uparrow$}} & \multicolumn{3}{c}{\textbf{CLIP Acc $\uparrow$}} \\
                           & & Top-1 & Top-3 & Top-5 \\
    \midrule
    COCO GT~\cite{(coco)lin2014microsoft}  & 67.07 & 0.7767 & 0.6236 & 0.5415 \\
    SAM~\cite{(SAM)kirillov2023segment}  & 68.17 & 0.7940 & 0.6521 & 0.5753 \\
    \textbf{ZIM (ours)} & \textbf{73.11} & \textbf{0.8616} & \textbf{0.7543} & \textbf{0.6855} \\
    \bottomrule
  \end{tabular}
  \end{adjustbox}
  \caption{
    \textbf{Quantitative results of image inpainting} using the Inpainting Anything framework~\cite{(inpaint_anything)yu2023inpaint}. The inpainting model takes three types of input masks (COCO ground-truth~\cite{(coco)lin2014microsoft}, SAM~\cite{(SAM)kirillov2023segment}, and ZIM) and we evaluate the corresponding inpainting results using CLIP distance and accuracy metrics~\cite{ekin2024clipaway, yildirim2023instinpaint}.
  }
  \label{tab:downstream_inpaint}
  \vspace{3mm}
\end{table}

\begin{figure*}[t]
    \centering
    \includegraphics[width=\linewidth]{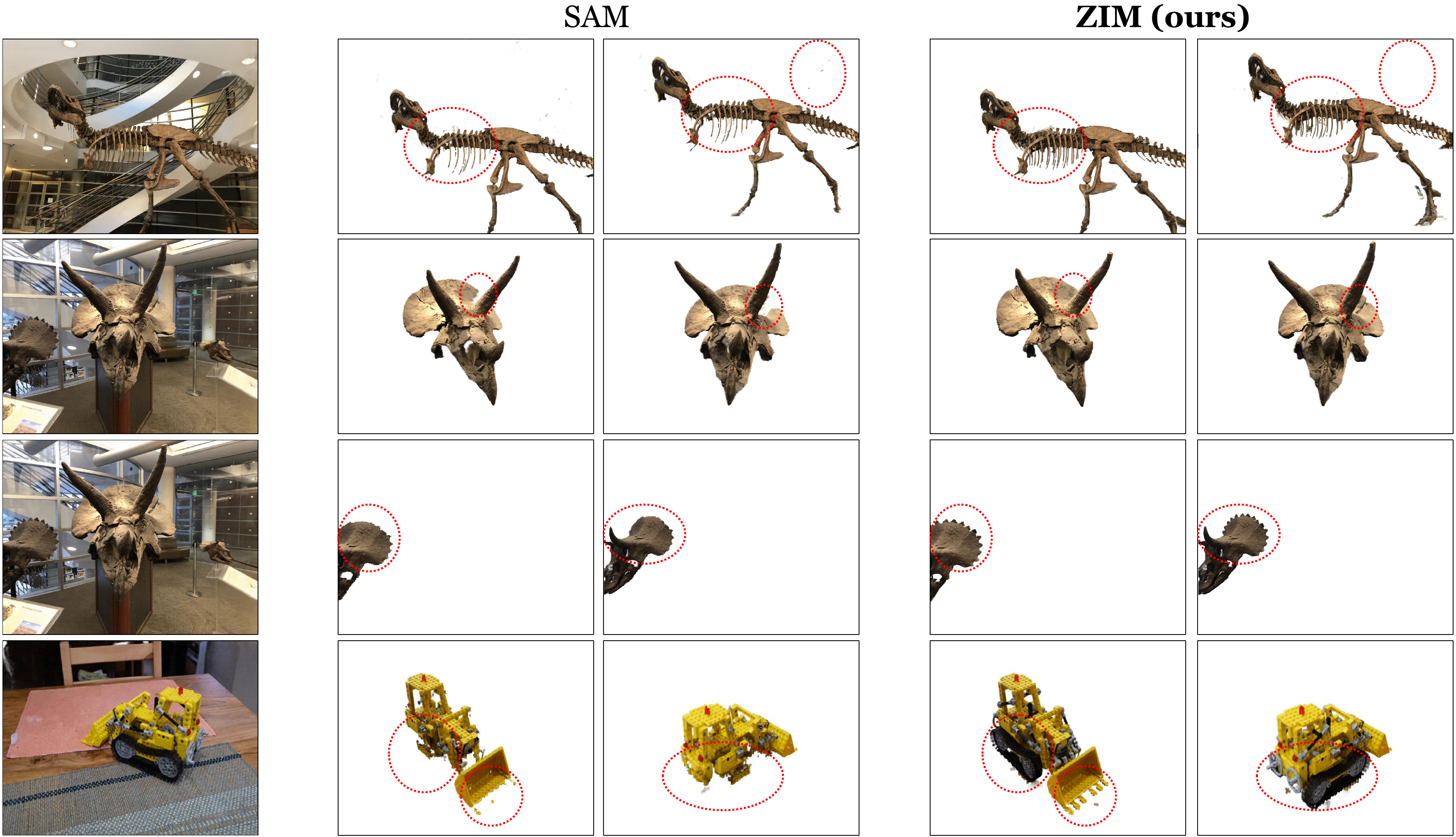}
    \caption{
    \textbf{Qualitative samples} of 3D object segmentation results guided by SAM~\cite{(SAM)kirillov2023segment} and ZIM models within the SA3D framework~\cite{(SA3D)cen2023segment} for the LLFF-trex, LLFF-horns~\cite{mildenhall2019llff}, and 360$^\circ$-kitchen~(Lego)~\cite{barron2022mipnerf360} datasets.
    }
    \label{fig:nerf}
    \vspace{7mm}
\end{figure*}

\subsection{3D Object Segmentation with NeRF}
The quality of the segmentation mask is crucial when converting a 2D mask into a 3D representation. In this work, we adopt the SA3D framework~\cite{(SA3D)cen2023segment}, which utilizes SAM to segment 3D objects from 2D masks by manually prompting the target object in a single view. By replacing SAM with ZIM in the 2D mask segmentation process, we significantly improve the quality of the resulting 3D objects. Figure \ref{fig:nerf} presents the qualitative results of segmented 3D objects guided by the SAM and ZIM models on the LLFF-trex and LLFF-horns~\cite{mildenhall2019llff} dataset. Compared to SAM, which often misses finer details due to its coarse-level mask generation, ZIM captures more intricate object features. These findings demonstrate that the precise matting capabilities of ZIM extend beyond 2D tasks, significantly enhancing the quality of 3D object segmentation.
For quantitative evaluation, we employ the NVOS~\cite{(NVOS)ren-cvpr2022-nvos} dataset, which includes finely annotated 2D masks. 
Since the SA3D framework relies on binary masks for projecting 2D masks into 3D space, we binarized the ZIM results using a threshold of 0.3. 
We fixed the number of self-prompting points at 10 and followed the experimental setup of SA3D~\cite{(SA3D)cen2023segment} for pre-trained NeRFs and manual prompts.
As shown in Table \ref{tab:downstream_nerf}, ZIM outperforms SAM in quantitative 3D segmentation results on the target view, reporting higher mask IoU scores. These findings demonstrate that the precise matting capabilities of ZIM extend beyond 2D tasks, significantly enhancing the quality of 3D object segmentation.

\begin{table}[t]
  \centering
  \begin{adjustbox}{max width=\linewidth}
  \begin{tabular}{l|cc}
    \toprule
    \multirow{2}{*}{Scenes} & \multicolumn{2}{c}{\textbf{Mask IoU (\%) $\uparrow$}} \\
                           & SAM & \textbf{ZIM} \\
    \midrule
    Fern & 84.9 & \textbf{86.6} \\
    Flower & 94.0 & \textbf{95.7} \\
    Fortress & 96.5 & \textbf{98.1} \\
    Horns-center & 94.0 & \textbf{97.4} \\
    Horns-left & 92.9 & \textbf{94.7} \\
    Leaves & 92.2 & \textbf{92.9} \\
    Orchids & 89.9 & \textbf{91.8} \\
    Trex & 83.7 & \textbf{85.4} \\
    \bottomrule
  \end{tabular}
  \end{adjustbox}
  \caption{
    \textbf{Quantitative results} of mask IoU scores on the target view for the NVOS dataset~\cite{(NVOS)ren-cvpr2022-nvos}.
  }
  \label{tab:downstream_nerf}
\end{table}

\subsection{Medical Image Segmentation}
Segmentation models are essential in medical image analysis, where they assist in identifying key anatomical structures and abnormalities. Building on the recent evaluation of SAM's performance in medical imaging by \cite{(medi_sam)mazurowski2023segment}, we explore the applicability of ZIM for zero-shot medical image segmentation. 
Given that neither SAM nor ZIM has been trained on medical image datasets, this experiment focuses on evaluating their zero-shot segmentation capabilities.
Both SAM and ZIM, using the ViT-B backbone, are evaluated across five medical imaging datasets: the hippocampus and spleen datasets from the Medical Image Decathlon~\cite{Antonelli_2022}, an ultrasonic kidney dataset~\cite{SONG2022106706}, an ultrasonic nerve dataset~\cite{ultrasound-nerve-segmentation}, and an X-ray hip dataset~\cite{x-ray}. 
Since these datasets comprise binary ground-truth masks, we apply a threshold of 0.3 to the matte output of ZIM.
Following the evaluation protocol from \cite{(medi_sam)mazurowski2023segment}, we employ five prompt modes: (1) a single point at the center of the largest contiguous region, (2) multiple points centered on up to three regions, (3) a box surrounding the largest region, (4) multiple boxes around up to three regions, and (5) a box encompassing the entire object.

\begin{figure}[t]
    \centering
    \includegraphics[width=0.95\linewidth]{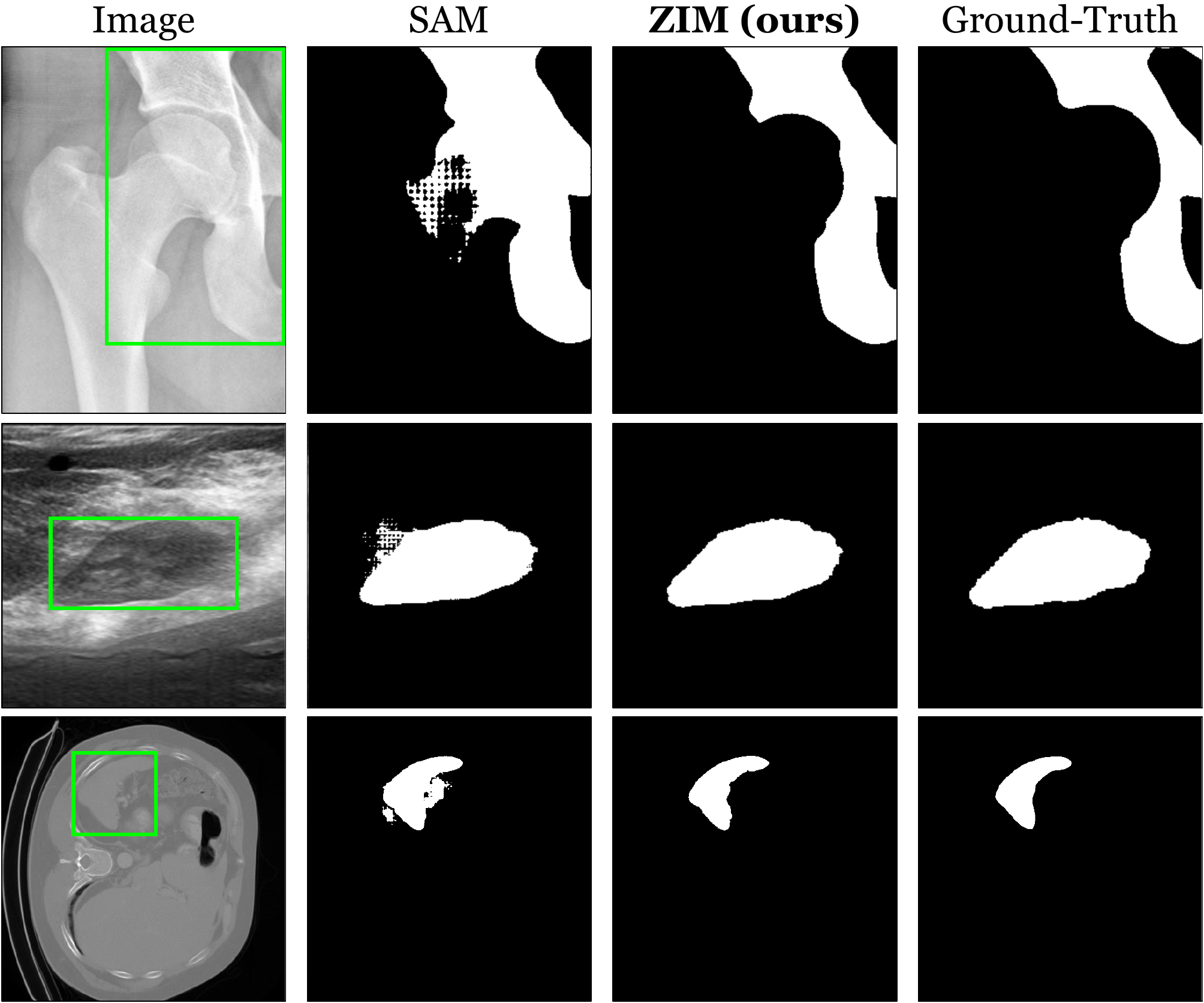}
    \caption{
    \textbf{Qualitative samples} of SAM and ZIM output masks on the medical image datasets~\cite{Antonelli_2022, SONG2022106706, ultrasound-nerve-segmentation, x-ray} using the box prompt.
    }
    \label{fig:medical_samples}
\end{figure}

\begin{figure}[t]
    \centering
    \includegraphics[width=\linewidth]{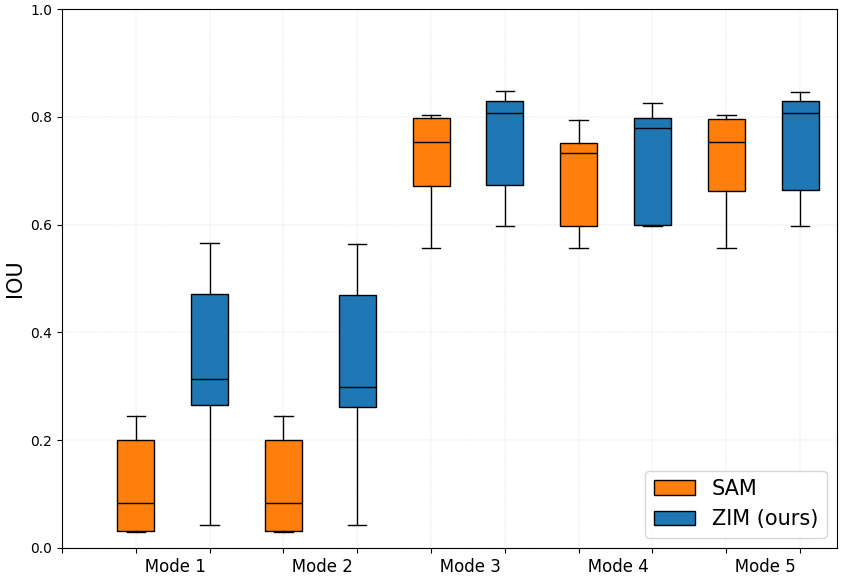}
    \caption{
    \textbf{Mask IoU distribution} across the five medical image analysis datasets~\cite{Antonelli_2022, SONG2022106706, ultrasound-nerve-segmentation, x-ray} for the five prompt modes.
    }
    \label{fig:medical_boxplot}
    \vspace{4mm}
\end{figure}

Figure \ref{fig:medical_boxplot} illustrates the distribution of IoU scores across the five datasets for each prompt mode. These results show that ZIM consistently outperforms SAM, particularly in point-based prompts (modes 1 and 2).
In addition, as shown in Figure \ref{fig:medical_samples}, SAM frequently exhibits checkerboard artifacts when dealing with the indistinct visual details of medical images. 
In contrast, ZIM produces more robust and precise segmentation masks due to our advanced pixel decoder. 
This demonstrates ZIM’s strong generalization on unseen and complex medical image data, highlighting its superior zero-shot capabilities compared to SAM.

\begin{table*}[t]
  \centering
  \begin{adjustbox}{max width=0.94\linewidth}
  \begin{tabular}{c|c|c|c|c|c|c|c|c|c|c|c}
    \toprule
    \multirow{2}{*}{Method} & \multirow{2}{*}{Backbone} & Latency & \multirow{2}{*}{Prompt} & \multicolumn{4}{c|}{\textbf{Fine-grained}{$\downarrow$}} & \multicolumn{4}{c}{\textbf{Coarse-grained}{$\downarrow$}} \\
     & & (ms){$\downarrow$} & & MSE & MSE$_{S}$ & MSE$_{M}$ & MSE$_{L}$ & MSE & MSE$_{S}$ & MSE$_{M}$ & MSE$_{L}$ \\
    \midrule
    \multirow{6}{*}{SAM~\cite{(SAM)kirillov2023segment}} & \multirow{2}{*}{ViT-B~\cite{(vit)dosovitskiy2020image}} & \multirow{2}{*}{172.5} & point & 21.651 & 2.717 & 7.145 & 70.502 & 5.569 & 4.092 & 5.405 & 77.761 \\
     & & & box & 11.057 & 0.329 & 2.983 & 38.501 & 1.044 & 0.181 & 2.456 & 24.519 \\
     \cmidrule{2-12}
      & \multirow{2}{*}{ViT-L~\cite{(vit)dosovitskiy2020image}} & \multirow{2}{*}{361.6} & point & 15.663 & 3.312 & 5.165 & 49.202 & 4.293 & 3.640 & 3.389 & 46.278 \\
     & & & box & 7.989 & 0.320 & 2.423 & 27.276 & 0.534 & \textbf{0.145} & 1.606 & 5.575 \\
     \cmidrule{2-12} 
      & \multirow{2}{*}{ViT-H~\cite{(vit)dosovitskiy2020image}} & \multirow{2}{*}{605.2} & point & 14.534 & 3.619 & 5.753 & 43.371 & 2.100 & \textbf{0.653} & 3.278 & 56.086 \\
     & & & box & 6.188 & 0.281 & 1.687 & 21.389 & 0.468 & 0.152 & 1.334 & 4.610 \\
    \midrule
    \multirow{4}{*}{SAM2~\cite{(SAM2)ravi2024sam}} & \multirow{2}{*}{Hiera-B+~\cite{(hiera)ryali2023hiera}} & \multirow{2}{*}{147.4} & point & 25.296 & 15.657 & 15.970 & 53.346 & 14.794 & 14.786 & 12.128 & 49.058 \\
     & & & box & 12.024 & 0.322 & 2.155 & 43.670 & 0.613 & 0.152 & 1.600 & 10.194 \\
     \cmidrule{2-12}
     & \multirow{2}{*}{Hiera-L~\cite{(hiera)ryali2023hiera}} & \multirow{2}{*}{195.0} & point & 16.937 & 0.871 & 5.613 & 56.702 & 2.572 & 0.954 & 4.308 & 57.804 \\
     & & & box & 10.616 & 0.263 & 1.888 & 38.636 & 0.704 & 0.149 & 1.424 & 18.019 \\
    \midrule
    \multirow{6}{*}{HQ-SAM~\cite{(HQ-SAM)ke2024segment}} & \multirow{2}{*}{ViT-B~\cite{(vit)dosovitskiy2020image}} & \multirow{2}{*}{177.2} & point & 36.674 & 5.094 & 9.356 & 123.199 & 6.457 & 3.602 & 9.596 & 102.834 \\
     & & & box & 42.457 & 0.392 & 4.369 & 160.456 & 2.733 & 0.230 & 2.521 & 124.372 \\
     \cmidrule{2-12}
     & \multirow{2}{*}{ViT-L~\cite{(vit)dosovitskiy2020image}} & \multirow{2}{*}{368.1} & point & 20.481 & 1.928 & 6.496 & 67.979 & 3.046 & 1.200 & 4.987 & 66.373 \\
     & & & box & 19.881 & 0.326 & 2.683 & 73.913 & 0.762 & 0.163 & 1.408 & 21.129 \\
     \cmidrule{2-12}
     & \multirow{2}{*}{ViT-H~\cite{(vit)dosovitskiy2020image}} & \multirow{2}{*}{608.1} & point & 22.547 & 5.308 & 6.601 & 71.446 & 3.599 & 1.659 & 5.034 & 77.643 \\
     & & & box & 23.743 & 0.263 & 2.538 & 89.518 & 0.789 & 0.164 & 1.266 & 24.469 \\
     \midrule
    \multirow{6}{*}{Matte-Any~\cite{(matte-any)yao2024matte}} & \multirow{2}{*}{ViT-B~\cite{(vit)dosovitskiy2020image}} & \multirow{2}{*}{668.5} & point & 20.844 & 3.381 & 7.953 & 65.108 & 6.053 & 4.739 & 5.897 & 70.443 \\
     & & & box & 9.746 & 0.918 & 3.856 & 31.109 & 1.983 & 1.235 & 3.039 & 24.426 \\
     \cmidrule{2-12}
     & \multirow{2}{*}{ViT-L~\cite{(vit)dosovitskiy2020image}} & \multirow{2}{*}{814.1} & point & 15.230 & 3.985 & 6.243 & 44.842 & 5.116 & 4.570 & 3.996 & 44.606 \\
     & & & box & 7.323 & 0.975 & 3.692 & 21.711 & 1.597 & 1.222 & 2.418 & 9.119 \\
     \cmidrule{2-12}
     & \multirow{2}{*}{ViT-H~\cite{(vit)dosovitskiy2020image}} & \multirow{2}{*}{1036.9} & point & 14.119 & 4.270 & 7.085 & 38.704 & 3.022 & 1.669 & 3.913 & 56.044 \\
     & & & box & 6.048 & 0.917 & 2.992 & 17.872 & 1.571 & 1.228 & 2.294 & 8.836 \\
     \midrule
    \multirow{2}{*}{Matting-Any~\cite{(matting-any)li2024matting}} & \multirow{2}{*}{ViT-B~\cite{(vit)dosovitskiy2020image}} & \multirow{2}{*}{200.3} & point & 77.335 & 27.256 & 41.349 & 202.692 & 36.187 & 31.549 & 40.993 & 197.340 \\
     & & & box & 68.372 & 18.286 & 33.111 & 192.566 & 23.780 & 17.344 & 36.194 & 175.418 \\
     \midrule
    \multirow{2}{*}{SMat~\cite{(smat)ye2024unifying}} & \multirow{2}{*}{ViT-B~\cite{(vit)dosovitskiy2020image}} & \multirow{2}{*}{263.4} & point & 123.664 & 31.549 & 40.993 & 197.340 & 59.113 & 52.305 & 68.709 & 250.587 \\
     & & & box & 133.515 & 17.344 & 36.194 & 175.418 & 61.157 & 53.280 & 74.099 & 261.544 \\
     \midrule
    \multirow{4}{*}{\textbf{ZIM (ours)}} & \multirow{2}{*}{ViT-B~\cite{(vit)dosovitskiy2020image}} & \multirow{2}{*}{187.8} & point & 8.213 & 0.870 & 3.962 & 24.934 & 1.788 & 1.444 & 1.731 & \textbf{18.983} \\
     & & & box & 1.893 & 0.205 & 2.228 & 3.617 & 0.448 & 0.200 & \textbf{1.382} & \textbf{0.632} \\
     \cmidrule{2-12}
     & \multirow{2}{*}{ViT-L~\cite{(vit)dosovitskiy2020image}} & \multirow{2}{*}{373.4} & point & \textbf{5.825} & \textbf{0.563} & \textbf{2.724} & \textbf{17.898} & \textbf{1.719} & \textbf{1.041} & \textbf{1.574} & 35.687 \\
     & & & box & \textbf{1.589} & \textbf{0.191} & \textbf{1.888} & \textbf{2.982} & \textbf{0.446} & \textbf{0.175} & 1.458 & 0.686 \\
    \bottomrule
  \end{tabular}
  \end{adjustbox}
  \vspace{-1mm}
  \caption{
    \textbf{Detailed Quantitative comparison} of our ZIM model and six existing methods on the MicroMat-3K dataset. 
    Results are presented for different backbone networks, model throughput, and MSE scores across object sizes (small, medium, and large).
    The latency is measured on the NVIDIA V100 GPU.
  }
  \label{tab:detail_analysis_mat}
  \vspace{-2mm}
\end{table*}

\section{Additional Experiemental Results}

\subsection{Detailed Experiment Settings}
We evaluate interactive matting performance using point and box prompts derived from ground-truth masks. For point sampling, we adopt RITM~\cite{(RITM)sofiiuk2022reviving} with a maximum of 12 points. Box prompts are extracted from the min-max foreground coordinates, with up to 10\% random perturbation to assess robustness to noisy box inputs. This sampling strategy is applied to MicroMat3K and public matting datasets such as AIM-500~\cite{(AIM-500)li2021deep} and P3M-500-NP~\cite{(P3M)li2021privacy}.

ZIM is trained on 1\% of SA1B-Matte, containing 2.2M matte labels. Since SAM pre-trained weights are utilized, increasing the training data to 10\% did not yield noticeable improvements, indicating that 1\% suffices for enhancing fine-grained representations.

All experiments are conducted using 8 V100 GPUs. Training the label converter and ZIM takes approximately 7 and 5 days, respectively.

\subsection{Detailed Quantitative Comparison}
Table \ref{tab:detail_analysis_mat} provides an in-depth evaluation of zero-shot matting models, with additional experiments utilizing various backbone networks: ViT-L and ViT-H~\cite{(vit)dosovitskiy2020image} for SAM~\cite{(SAM)kirillov2023segment}, HQ-SAM~\cite{(HQ-SAM)ke2024segment}, and Matte-Any~\cite{(matte-any)yao2024matte}, and Hiera-L~\cite{(hiera)ryali2023hiera} for SAM2~\cite{(SAM2)ravi2024sam}.
Furthermore, we investigate model behavior based on the size of the target object by defining three object size groups, according to the ratio of the foreground region in the image: small (ratio $<$ 1\%), medium (1\% $\leq$ ratio $<$ 10\%), and large (ratio $\geq$ 10\%). The MSE error is reported for each object size group, offering a detailed understanding of model performance across varying object sizes. Additionally, we measure the model throughput using an NVIDIA V100 GPU to assess computational efficiency.

The results in Table \ref{tab:detail_analysis_mat} provide some meaningful insights:
(1) ZIM consistently outperforms SAM, especially for larger objects, as indicated by the $MSE_{L}$ metric. This improvement is likely due to the reduction of checkerboard artifacts, a known issue in SAM’s pixel decoder, which our advanced decoder addresses effectively, as evidenced in Figures 1, \ref{fig:amg_1}, and \ref{fig:amg_2}.
(2) ZIM demonstrates highly competitive results even with the smaller ViT-B backbone, outperforming models like SAM and HQ-SAM with larger backbones such as ViT-H. Additionally, the performance of ZIM with the ViT-L backbone suggests that further improvements could be achieved with more powerful architectures.
(3) Despite our advanced decoder, ZIM introduces only a marginal increase in latency (just 10ms more than SAM) making it a lightweight and efficient option for zero-shot matting tasks.
(4) Compared to existing matting models ($e.g.,$ Matte-Any, Matting-Any, and SMat), ZIM delivers superior performance while maintaining efficiency.

\subsection{Discussion on Transparency Prediction}
In the image matting task, there is a distinct scenario, that is, predicting the transparency of objects, such as glasses and fire.
Due to its inherent difficulties, existing matting methods~\cite{(indexmat)lu2019indices,(MGM)(RWP-636)yu2021mask,(trans460)cai2022transmatting} typically rely on curated transparency datasets~\cite{(trans460)cai2022transmatting} containing close-up transparent object images with clear backgrounds, making them effective in constrained environments but less adaptable to open-world scenarios. ZIM, on the other hand, is designed for fine-grained mask representation in general object segmentation across diverse and complex scenes. However, predicting transparency within ZIM remains a challenge due to the lack of large-scale open-world transparency datasets and the inherent complexity of identifying transparent regions in natural images.

\begin{table}[t]
  \centering
  \begin{adjustbox}{max width=0.95\linewidth}
  \begin{tabular}{c|c|cc}
    \toprule
    \multirow{2}{*}{Model} & \multirow{2}{*}{Input} & \multicolumn{2}{c}{\textbf{Transparent-460}~\cite{(trans460)cai2022transmatting}} \\
                           & & MSE $\downarrow$ & SAD $\downarrow$  \\
    \midrule
    IndexNet~\cite{(indexmat)lu2019indices} & Trimap & 112.53 & 573.09 \\
    MGMatting~\cite{(MGM)(RWP-636)yu2021mask} & Trimap & 6.33 & 111.92 \\
    TransMatting~\cite{(trans460)cai2022transmatting} & Trimap & 4.02 & 88.34 \\
    \textbf{ZIM (ours)} & Box & 15.55 & 298.78 \\
    \bottomrule
  \end{tabular}
  \end{adjustbox}
  \caption{
    \textbf{Quantitative results on the transparent object matting dataset, Transprent-460~\cite{(trans460)cai2022transmatting}}.
  }
  \label{tab:comparison_trans}
  \vspace{-5mm}
\end{table}

To investigate ZIM’s adaptability to transparency prediction, we fine-tune it on the Transparent-460 dataset~\cite{(trans460)cai2022transmatting} and compare its performance against specialized transparency matting methods~\cite{(indexmat)lu2019indices, (MGM)(RWP-636)yu2021mask, (trans460)cai2022transmatting}, as shown in Table \ref{tab:comparison_trans}. While these methods leverage trimaps to provide explicit spatial guidance for transparency estimation, ZIM relies only on sparse box prompts, which offer less precise object boundary information. Despite this limitation, ZIM achieves reasonable results, demonstrating its strong transferability.

\subsection{Expanding Prompt Sources}
Interactive models, such as SAM, commonly support only point and box prompts, which can be limiting for complex scenarios requiring more intuitive user interaction.
Here, we demonstrate that ZIM offers a more flexible approach by expanding the variety of prompt types, including text and scribble prompts, enabling more versatile and user-friendly matting applications.

\begin{figure}[t]
    \centering
    \includegraphics[width=\linewidth]{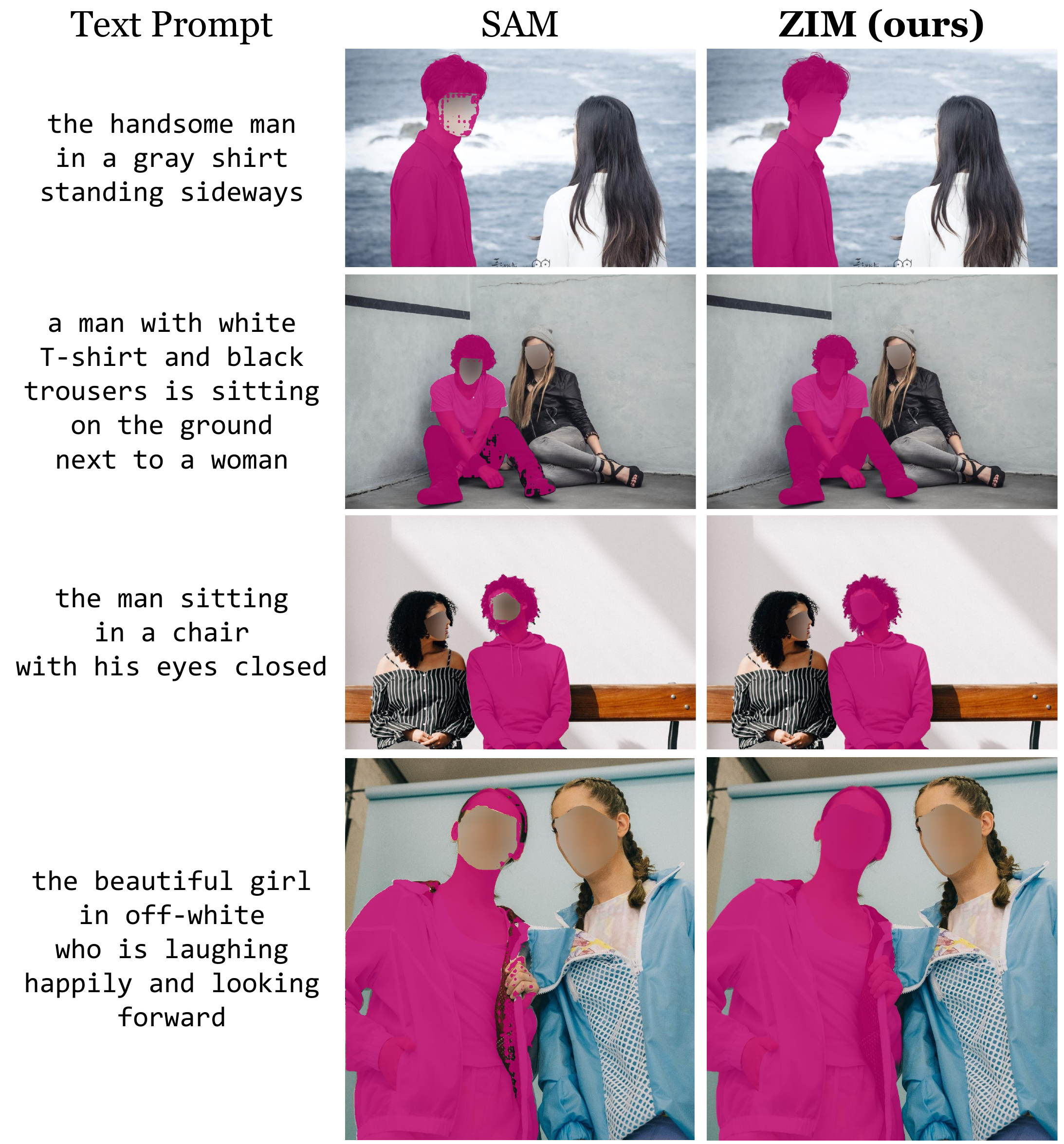}
    \caption{
    \textbf{Qualitative samples} of text prompting results using Grounded-ZIM on the RefMatte-RW100~\cite{(RefMatte)li2023referring} dataset. Grounded-ZIM replaces SAM with ZIM in the Grounded-SAM framework~\cite{(grounded_sam)ren2024grounded}, leveraging Grounding DINO~\cite{(GDINO)liu2024grounding}'s open-set object detection with ZIM's superior matting capabilities to produce more precise and robust results than the original Grounded-SAM.
    }
    \label{fig:text_prompt}
\end{figure}

\paragraph{Text Prompt.}
To enable text prompts, we integrate ZIM with the Grounded-SAM framework~\cite{(grounded_sam)ren2024grounded}.
Grounded-SAM combines Grounding DINO~\cite{(GDINO)liu2024grounding}, an open-set object detector, with the Segment Anything Model (SAM) to process image-text pairs and return bounding boxes for objects mentioned in the text.
In our approach, we replace SAM with ZIM, constructing Grounded-ZIM that leverages Grounding DINO's robust detection capabilities while utilizing ZIM's superior matting performance for fine-grained boundary refinement.
This combination enables natural language-driven matting without requiring precise manual annotation.
ZIM then uses the detected bounding boxes as prompts to produce detailed matte outputs with enhanced edge quality and alpha channel precision.
As shown in Figure \ref{fig:text_prompt}, Grounded-ZIM provides high-quality outputs with a simple text prompting pipeline, offering more precise and robust mask generation than the original Grounded-SAM framework.

\begin{figure}[!t]
    \centering
    \includegraphics[width=\columnwidth]{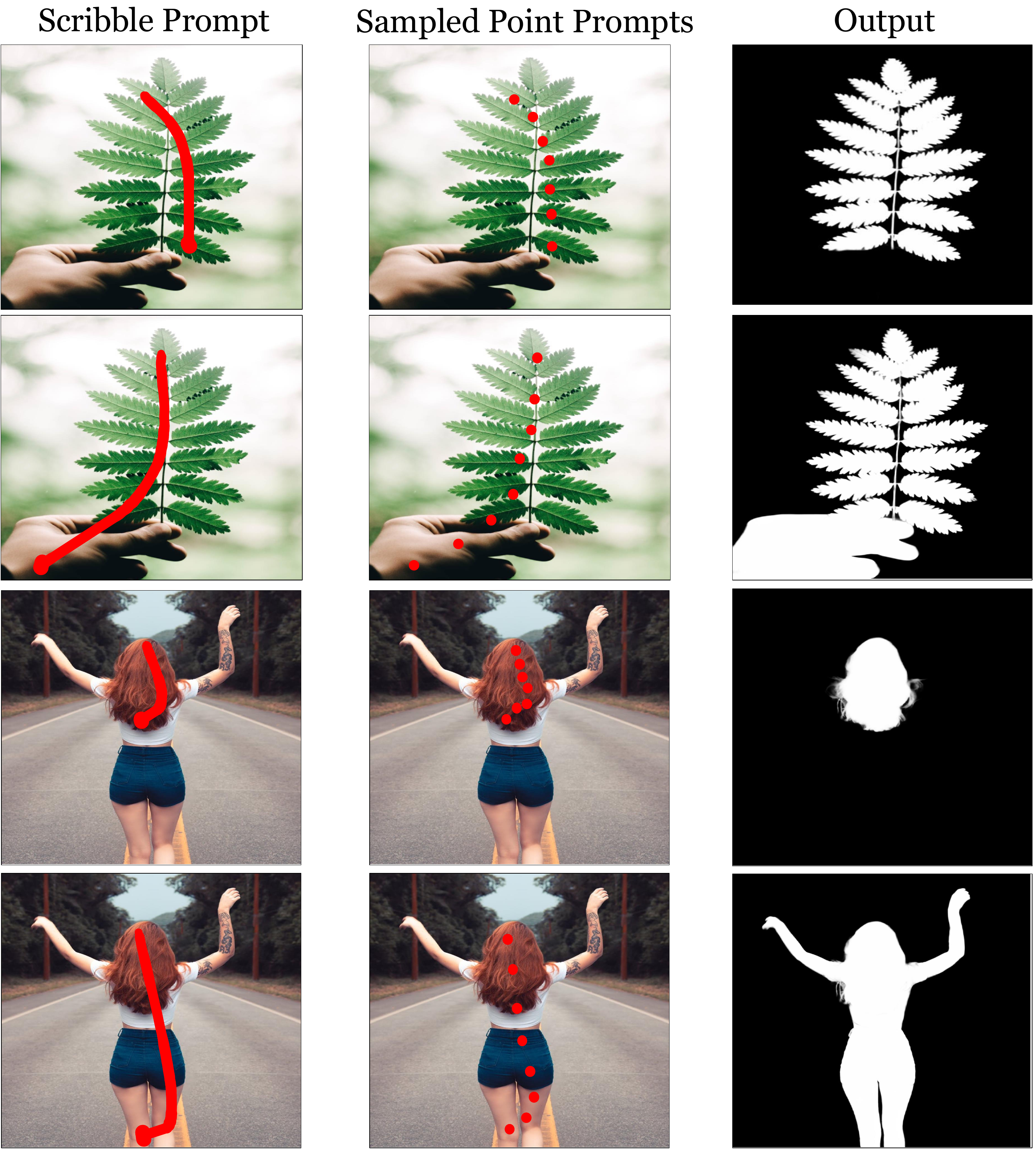}
    \caption{
    \textbf{Qualitative examples} of scribble prompting results. ZIM processes scribble inputs through uniform point sampling (max 24 points) along the scribble path, generating accurate and stable matte outputs for marked regions of interest.
    }
    \label{fig:scribble}
\end{figure}

\paragraph{Scribble Prompt.}
In addition, ZIM can support scribble prompts, which provide users with an intuitive way to mark regions of interest without requiring precise boundary delineation.
We implement this functionality by sampling points along the scribble path, converting the continuous stroke into discrete point prompts that ZIM can process effectively.
To ensure comprehensive coverage of the scribble region while maintaining computational efficiency, we employ uniform sampling with the maximum number of sampled points set to 24.
This design strikes a balance between prompt coverage and processing speed, allowing ZIM to effectively handle the scribble input and generate high-quality matte outputs with smooth boundaries.
Figure \ref{fig:scribble} shows an example of how the scribble prompt leads to accurate and stable results across diverse scenarios.
These expansions highlight the versatility of ZIM in accommodating diverse input modalities, making it suitable for various real-world applications requiring different levels of user interaction.

\subsection{Additional Ablation Study}
\paragraph{Impact of our proposed components on existing segmentation models.}
Table \ref{tab:additional_analysis} demonstrates that both the Label Converter and SA1B-Matte training data provide substantial improvements to baseline performance on the MicroMat3K fine-grained dataset.
The Label Converter, when applied to SAM's coarse output masks, reduces MSE from 11.05 to 7.34, representing a significant improvement in fine-grained mask quality. However, the converter cannot address inherent limitations in SAM's base predictions, such as checkerboard artifacts, resulting in performance that remains below ZIM's 1.89 MSE.

In addition, training existing models ($i.e.,$ SAM, HQ-SAM, and Matting-Any) on our SA1B-Matte dataset yields consistent improvements across all models. SAM trained with SA1B-Matte achieves 2.71 MSE, while HQSAM and Matting-Any reach 13.35 and 12.26 MSE, respectively. Despite these improvements, all existing models underperform compared to ZIM, highlighting the importance of our architectural innovations, including the Hierarchical Pixel Decoder and Prompt-Aware Masked Attention. The performance gap is particularly notable for HQ-SAM and Matting-Any, which rely on the frozen SAM model that limits their capacity to learn fine-grained representations during training.

\begin{table}[t]
  \centering
  \begin{adjustbox}{max width=\linewidth}
  \begin{tabular}{c|c}
    \toprule
    Model & MSE{$\downarrow$} \\
    \midrule
    SAM~\cite{(SAM)kirillov2023segment} & 11.05 \\
    SAM~\cite{(SAM)kirillov2023segment} + Converter & 7.34 \\
    SAM~\cite{(SAM)kirillov2023segment} + SA1B-Matte & 2.71 \\
    \midrule
    HQSAM~\cite{(HQ-SAM)ke2024segment} & 42.45 \\
    HQSAM~\cite{(HQ-SAM)ke2024segment} + SA1B-Matte & 13.35 \\
    \midrule
    Matting-Any~\cite{(matting-any)li2024matting} & 68.37 \\
    Matting-Any~\cite{(matting-any)li2024matting} + SA1B-Matte & 12.26 \\
    \midrule
    \textbf{ZIM (ours)} & \textbf{1.89} \\
    \bottomrule
  \end{tabular}
  \end{adjustbox}
  \caption{
    \textbf{Analysis} of our proposed components ($i.e.,$ Label Converter and SA1B-Matte dataset) when applied to existing models.
  }
  \label{tab:additional_analysis}
\end{table}

\paragraph{Effect of Hyperparameter $\sigma$.}
The hyperparameter $\sigma$ controls the standard deviation of the 2D Gaussian map used to create the soft attention mask for point prompts. 
A larger $\sigma$ results in a wider spread of the Gaussian, covering a broader region around the point. 
Table \ref{tab:analysis_sigma} presents the performance of the ZIM model using point prompts on the MicroMat-3K dataset for varying values of $\sigma$. 
Through experimentation, we found that setting $\sigma$ to 21 strikes a balance by generating an appropriately sized soft attention mask that effectively captures relevant features while minimizing unnecessary coverage.

\paragraph{Effect of Hyperparameter $\lambda$.}
The hyperparameter $\lambda$, as defined in Eq (1), controls the weight assigned to the Gradient loss, influencing the emphasis on edge detail during training. 
We evaluate how different values of $\lambda$ affect the performance of the ZIM on the MicroMat-3K dataset. 
The results in Table \ref{tab:analysis_lambda} indicate that a $\lambda$ value of 10 achieves optimal performance, providing a balanced trade-off between smoothness and edge accuracy.

\subsection{Additional Qualitative Samples}
\begin{figure*}
    \centering
    \includegraphics[width=0.8\linewidth]{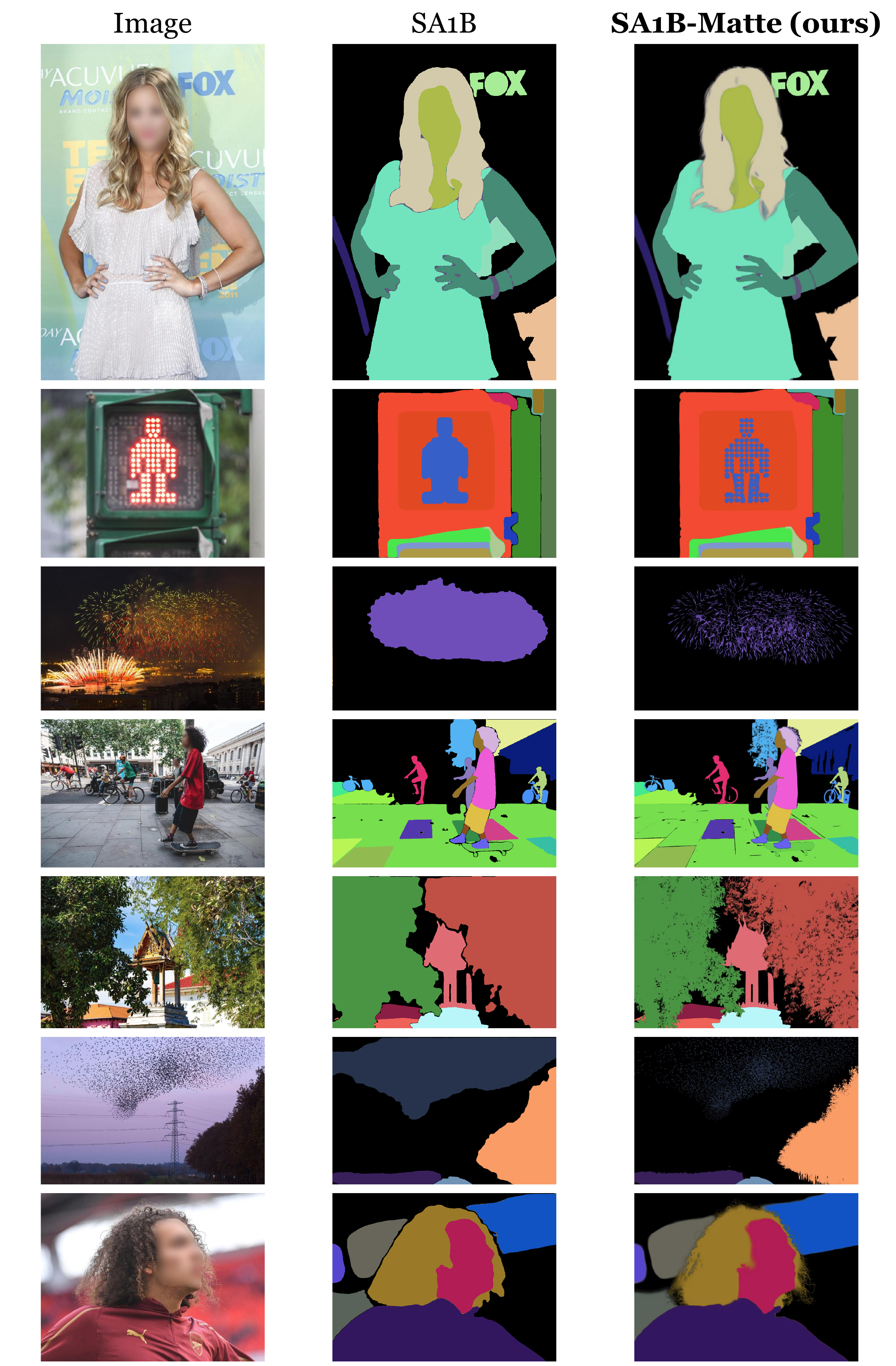}
    \caption{
    \textbf{Additional qualitative samples} of the SA1B dataset~\cite{(SAM)kirillov2023segment} with micro-level coarse labels and our SA1B-Matte dataset with micro-level fine labels.
    }
    \label{fig:appendix_dataset}
\end{figure*}

\begin{figure*}[t]
    \centering
    \begin{subfigure}[b]{0.95\linewidth}
        \centering
        \includegraphics[width=0.93\linewidth]{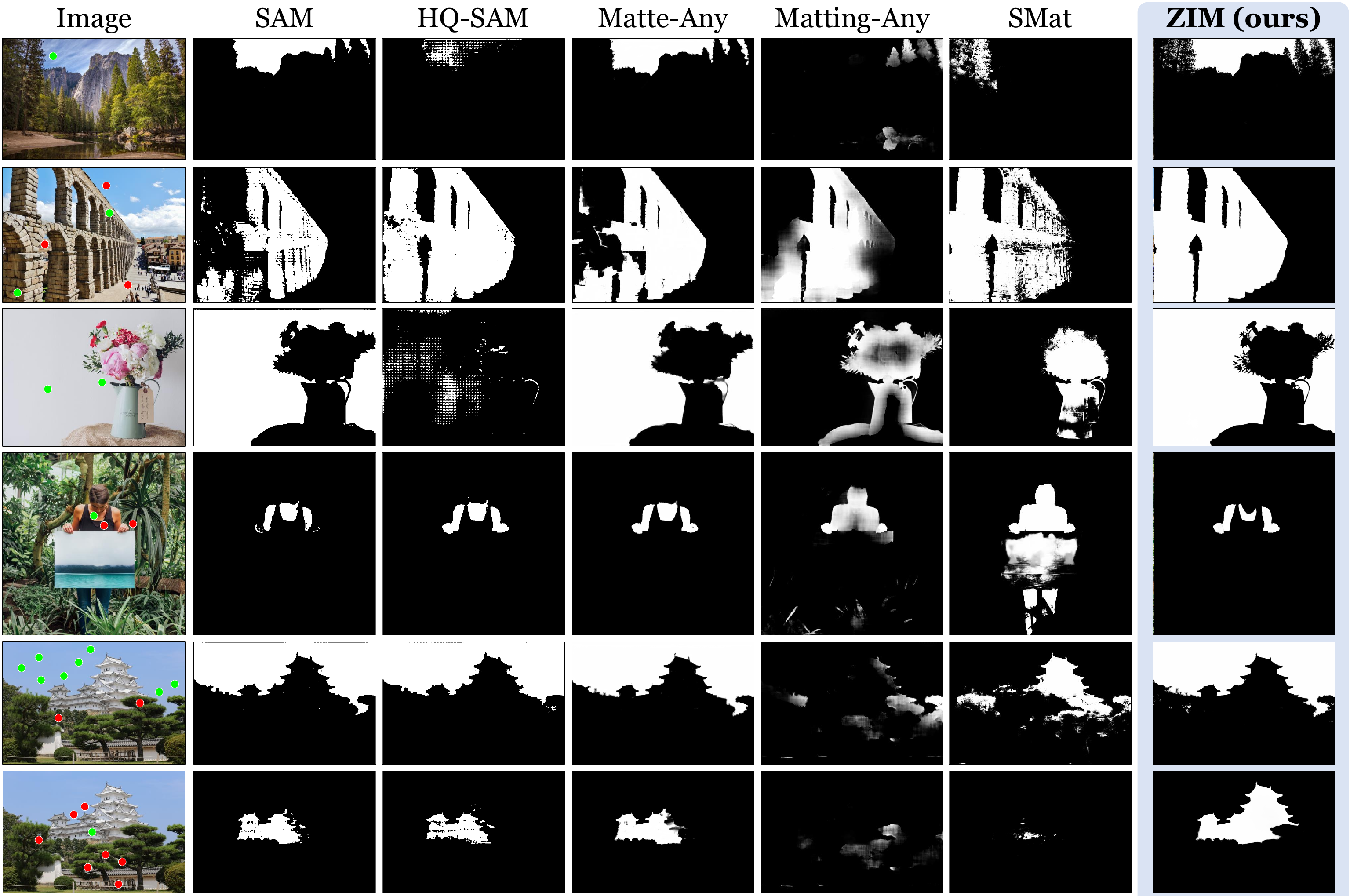} 
        \caption{}
        \label{fig:appendix_output_point}  
        \vspace{3mm}
    \end{subfigure}
    \begin{subfigure}[b]{0.95\linewidth}
        \centering
        \includegraphics[width=0.93\linewidth]{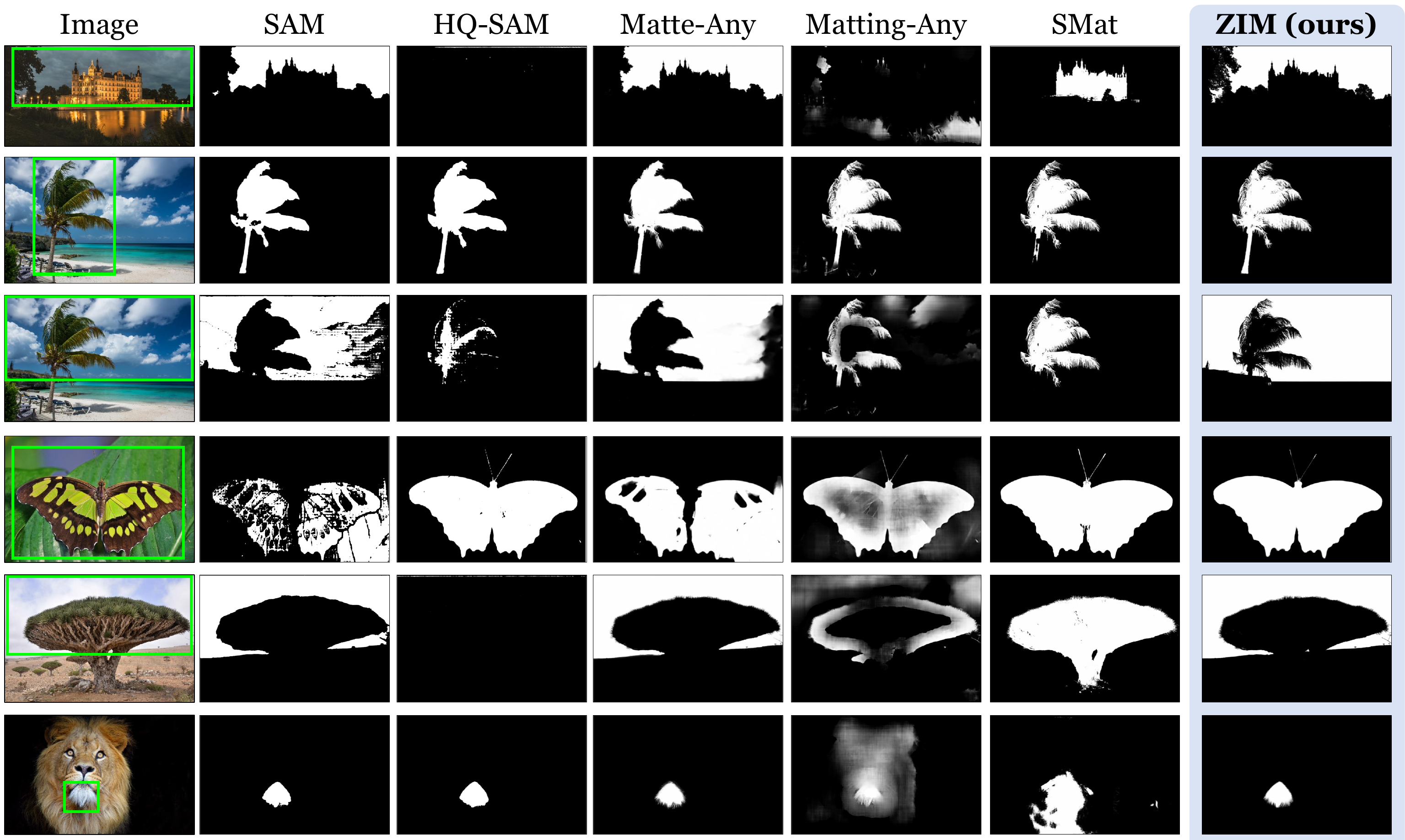} 
        \caption{}
        \label{fig:appendix_output_box} 
    \end{subfigure}
    \caption{
        \textbf{Additional qualitative samples} of ZIM with five existing zero-shot models (SAM~\cite{(SAM)kirillov2023segment}, HQ-SAM~\cite{(HQ-SAM)ke2024segment}, Matte-Any~\cite{(matte-any)yao2024matte}, Matting-Any~\cite{(matting-any)li2024matting}, and SMat~\cite{(smat)ye2024unifying}) based on (a) point prompts and (b) box prompts. 
    }
    \label{fig:appendix_output}
\end{figure*}

We provide more qualitative samples for the SA1B-Matte dataset (Figure \ref{fig:appendix_dataset}) and ZIM output mattes (Figure \ref{fig:appendix_output}).

\begin{table}[t]
    \centering
    \begin{subtable}{\linewidth}
        \centering
        \begin{adjustbox}{max width=\linewidth}
          \begin{tabular}{c|c|c|c|c|c|c}
            \toprule
            \multirow{2}{*}{$\sigma$} & \multicolumn{3}{c|}{\textbf{Fine-grained}} & \multicolumn{3}{c}{\textbf{Coarse-grained}} \\
             & SAD{$\downarrow$} & MSE{$\downarrow$} & Grad{$\downarrow$} & SAD{$\downarrow$} & MSE{$\downarrow$} & Grad{$\downarrow$} \\
            \midrule
            11 & 31.476 & 8.381 & 5.505 & 6.741 & 1.816 & 1.538 \\
            \textbf{21} & \textbf{31.286} & \textbf{8.213} & \textbf{5.324} & \textbf{6.645} & \textbf{1.788} & \textbf{1.469} \\
            41 & 31.341 & 8.298 & 5.476 & 6.686 & 1.807 & 1.493 \\
            \bottomrule
          \end{tabular}
          \end{adjustbox}
      \caption{}
      \label{tab:analysis_sigma}
    \end{subtable}%
    \vspace{2mm}
    \begin{subtable}{\linewidth}
        \centering
        \begin{adjustbox}{max width=\linewidth}
        \begin{tabular}{c|c|c|c|c|c|c}
            \toprule
            \multirow{2}{*}{$\lambda$} & \multicolumn{3}{c|}{\textbf{Fine-grained}} & \multicolumn{3}{c}{\textbf{Coarse-grained}} \\
             & SAD{$\downarrow$} & MSE{$\downarrow$} & Grad{$\downarrow$} & SAD{$\downarrow$} & MSE{$\downarrow$} & Grad{$\downarrow$} \\
            \midrule
            5 & 33.086 & 9.056 & 7.392 & 6.966 & 1.970 & 1.817 \\
            \textbf{10} & \textbf{31.286} & \textbf{8.213} & \textbf{5.324} & \textbf{6.645} & \textbf{1.788} & \textbf{1.469} \\
            20 & 31.402 & 8.295 & 5.356 & 6.712 & 1.838 & 1.475 \\
            \bottomrule
          \end{tabular}
          \end{adjustbox}
      \caption{}
      \label{tab:analysis_lambda}
    \end{subtable}
    \label{tab:analysis_additional}
    \vspace{-6mm}
    \caption{\textbf{Analysis} of ZIM using point prompt evaluations: 
    (a) Effect of the hyperparameter $\sigma$ and (b) Effect of the hyperparameter $\lambda$.
    }
\end{table}


\end{document}